\pdfoutput=1

\documentclass[twoside]{article}

\usepackage[accepted]{aistats2019}
\usepackage[round]{natbib}

\usepackage{tikz}
\usepackage{enumitem}
\usepackage{algorithmicx}
\usepackage{algorithm}
\usepackage{algpseudocode}
\usepackage{cases}
\usepackage{math-common}

\begin{document}

%
\runningtitle{Clustering Time Series with Nonlinear Dynamics}

%
\runningauthor{A. Lin, Y. Zhang, J. Heng, S. A. Allsop, K. M. Tye, P. E. Jacob, D. Ba}

\twocolumn[

\aistatstitle{Clustering Time Series with Nonlinear Dynamics: A Bayesian Non-Parametric and Particle-Based Approach}

\aistatsauthor{ Alexander Lin$^*$ \And Yingzhuo Zhang$^*$ \And Jeremy Heng$^{*\dag}$ \And Stephen A. Allsop$^\ddag$}
\vspace{-0.1em}
\aistatsauthor{ Kay M. Tye$^\S$ \And Pierre E. Jacob$^*$ \And Demba Ba$^*$ }
 \vspace{-0.1em}
\aistatsaddress{ $^*$Harvard University \And $^\dag$ESSEC Business School \And $^\ddag$MIT, Picower Institute \And $^\S$Salk Institute} 
]

\begin{abstract}
We propose a general statistical framework for clustering multiple time series that exhibit nonlinear dynamics into an a-priori-unknown number of sub-groups. Our motivation comes from neuroscience, where an important problem is to identify, within a large assembly of neurons, subsets that respond similarly to a stimulus or contingency.  Upon modeling the multiple time series as the output of a Dirichlet process mixture of nonlinear state-space models, we derive a Metropolis-within-Gibbs algorithm for full Bayesian inference that alternates between sampling cluster assignments and sampling parameter values that form the basis of the clustering. The Metropolis step employs recent innovations in particle-based methods.  We apply the framework to clustering time series acquired from the prefrontal cortex of mice in an experiment designed to characterize the neural underpinnings of fear.
\end{abstract}

\section{INTRODUCTION}

In a data set comprising hundreds to thousands of neuronal time series~\citep{brown2004multiple}, the ability to automatically identify sub-groups of time series that respond similarly to an exogenous stimulus or contingency can provide insights into how neural computation is implemented at the level of groups of neurons. 


Existing methods for clustering multiple time series can be classified broadly into feature-based approaches and model-based ones. The former extract a set of features  from each time series, followed by clustering in feature space using standard algorithms, e.g. \cite{humphries2011spike}. While simple to implement, feature-based approaches cannot be used to perform statistical inference on the parameters of a physical model by which the time series are generated. 

Previous model-based approaches for clustering multiple time series typically employ Bayesian mixtures of time series models.  Examples have included GARCH models \citep{bauwens2007bayesian}, INAR models \citep{roick2019clustering}, and TRCRP models \citep{saad2018temporally}.  


State-space models are a well-known, flexible class of models for time series data \citep{durbin2012time}.  Many existing model-based approaches for clustering time series use a mixture of linear Gaussian state-space models. 
~\cite{inoue2006cluster} and~\cite{chiappa2007output} both consider the case of finite mixtures and use Gibbs sampling and variational-Bayes respectively for posterior inference.~\cite{nieto2014bayesian} and~\cite{middleton2014} use a Dirichlet process mixture to infer the number of clusters and Gibbs sampling for full posterior inference.
In all cases, the linear-Gaussian assumption is crucial; it enables exact evaluation of the likelihood using a Kalman filter and the ability to sample exactly from the state sequences underlying each of the time series. For nonlinear and non-Gaussian state-space models, this likelihood cannot be evaluated in closed form and exact sampling is not possible. 

We introduce a framework for clustering multiple time series that exhibit nonlinear dynamics into an a-priori-unknown number of clusters, each modeled as a nonlinear state-space model.  We derive a Metropolis-within-Gibbs algorithm for inference in a Dirichlet process mixture of state-space models with linear-Gaussian states and binomial observations, a popular model in the analysis of neural spiking activity~\citep{smith2003estimating}. The Metropolis step uses particle marginal Metropolis Hastings~\citep{andrieu2010particle}, which requires likelihood estimates with small variance. We use controlled sequential Monte Carlo~\citep{heng2017controlled} to produce such estimates. We apply the framework to the clustering of 33 neural spiking time series acquired from the prefrontal cortex of mice in an experiment designed to characterize the neural underpinnings of fear. The framework produces a clustering of the neurons into groups that represent {various degrees of neuronal signal modulation.}

\section{NONLINEAR TIME SERIES CLUSTERING MODEL}

We begin by introducing the Dirichlet Process nonlinear State-Space Mixture (DPnSSM) model for clustering multiple time series exhibiting nonlinear dynamics.

\subsection{DPnSSM} \label{sec:DPnSSM}

Let $\bd Y = \{\bidx y 1, \ldots, \bidx y N\}$ be a set of observed time series in which each series $\bidx y n = \idx[1] y n, \ldots, \idx[T] y n$ is a vector of length $T$.  Following the framework of state-space models, we model $\bidx y n$ as the output of a latent, autoregressive process $\bidx x n$.  For all $n = 1, \ldots, N$,
\begin{align}
{\idx[1] x n} \given \idx {\bd{\tilde{\theta}}} n &\sim h(\idx[1] x n; \idx {\bd{\tilde{\theta}}} n), \\
\idx[t] x n \given \idx[t-1] x n, \idx {\bd{\tilde{\theta}}} n &\sim f(\idx[t-1] x n, \idx[t] x n; \idx {\bd{\tilde{\theta}}} n), &   1 < t,  \nonumber \\ 
\idx[t] y n \given \idx[t] x n, \idx {\bd{\tilde{\theta}}} n &\sim g(\idx[t] x n, \idx[t] y n; \idx {\bd{\tilde{\theta}}} n ),  &  1 \leq t, \nonumber
\end{align}
where $\idx {\bd{\tilde{\theta}}} n$ denotes a set of \emph{parameters} for series $n$, $f$ is some \emph{state transition density}, $g$ is some \emph{state-dependent likelihood}, and {$h$ is some \emph{initial prior}}.  

The hidden parameters ${\bd{\tilde{\Theta}}}  = \{\idx {\bd{\tilde{\theta}}}  1, \ldots, \idx {\bd{\tilde{\theta}}}  N\}$ form the basis of clustering $\bd Y$; that is, if $\bidx \ttheta n = \bidx \ttheta {n'}$, then series $n$ and $n'$ belong to the same cluster.  However, in many applications, the number of clusters itself may be unknown and, therefore, we choose to model the parameters as coming from a distribution $Q$ that is sampled from a Dirichlet process (DP) with base distribution $G$ and inverse-variance parameter $\alpha$.  \cite{ferguson1973bayesian} showed that $Q$ is almost surely discrete and that the number of distinct values within $N \to \infty$ draws from $Q$ is random.  Thus, the DP serves as a prior for discrete distributions over elements in the support of $G$.  The overall objective is to infer the joint distribution of ${\bd{\tilde{\Theta}}} \given \bd Y, \alpha, G$.

The Chinese Restaurant Process (CRP) representation of the DP integrates out the intermediary distribution $Q$ used to generate ${\bd{\tilde{\Theta}}}$ \citep{neal2000markov}.  The CRP allows us to nicely separate the process of assigning a cluster (i.e. table) to each $\bidx y n$ from the process of choosing a hidden parameter (i.e. table value) for each cluster.  This is similar to the finite mixture model, but we do not need to choose $K$, the number of clusters, a priori.

\begin{figure}[h]
\begin{center}
\includegraphics[scale=0.4]{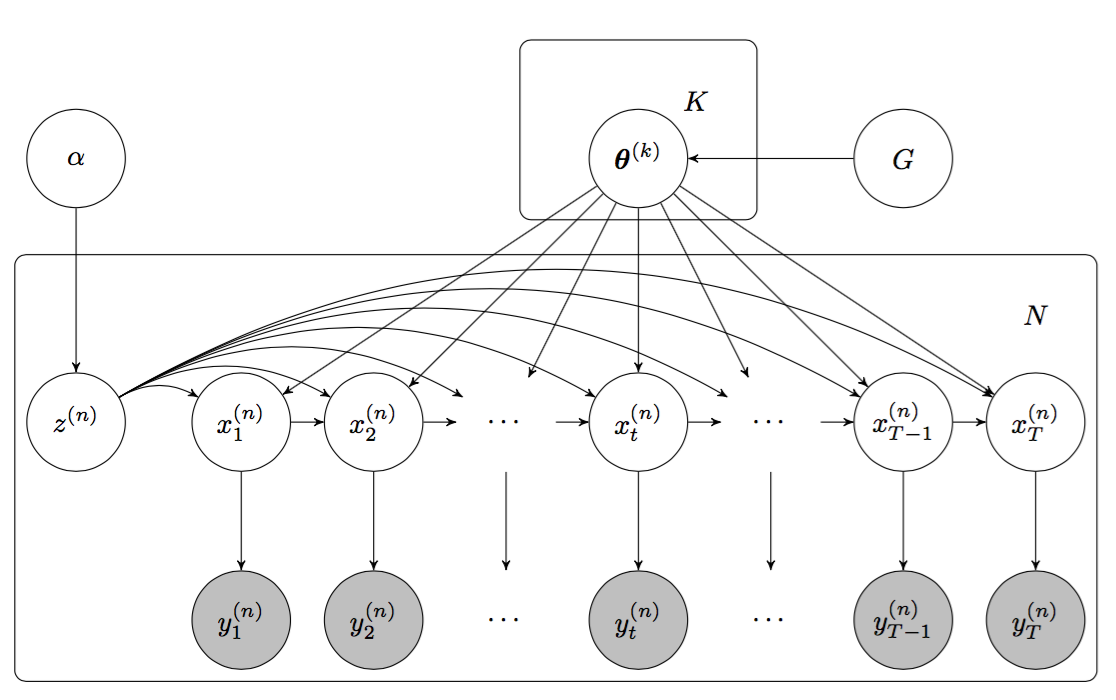}
\end{center}
\caption{{The graphical model representation of the DPnSSM. Observations are shown in grey, and states and parameters are shown in white. For simplicity, we omit the dependencies that are assumed in the DPnSSM between $\bd Y$ and ($Z$, $\bd \Theta$).}}
\label{model-graph}
\end{figure}

Under the CRP, we index the parameters by the cluster index $k$ instead of the observation index $n$.  Let $\idx z n \in \{1, \ldots, K\}$ denote the cluster identity of series $n$ and let $\idx {\bd{\theta}}  k$ denote the hidden parameters for cluster $k$. We formally define the model as follows,
\begin{align} \label{crp}
\idx z 1, \ldots, \idx z N, K \given \alpha &\sim \text{CRP}(\alpha, N), \\ 
\idx {\bd{\theta}}  k \given G &\sim G, & \quad 1 \leq k \leq K, \nonumber 
\end{align}
and for all $n = 1,\ldots,N$, 
\begin{align}
{\idx[1] x n} \given \idx z n = k, \idx {\bd{\theta}} k & \sim h(\idx[1] x n; \idx {\bd{\theta}} k), \nonumber \\ 
\idx[t] x n \given \idx[t-1] x n, \idx z n = k, \idx {\bd{\theta}} k &\sim f(\idx[t-1] x n, \idx[t] x n; \idx {\bd{\theta}}  k), & 1 < t, \nonumber \\  
\idx[t] y n \given \idx[t] x n, \idx z n = k, \idx {\bd{\theta}} k &\sim g(\idx[t] x n, \idx[t] y n; \idx {\bd{\theta}}  k), & 1 \leq t. \nonumber
\end{align}   
Let $Z = \{z^{(1)},\ldots,z^{(N)}\}$, $\bd \Theta = \{\bd \theta^{(1)},\ldots,\bd \theta^{(K)}\}$.  Figure \ref{model-graph} shows the graphical model for the DPnSSM.  Extensions of the model and inference algorithm to handling multi-dimensional time series and finite mixtures can be found in \textbf{Appendix \ref{model-extensions}}.

\subsection{Point Process State-Space Model}
\label{ppssm}
While the DPnSSM is defined for generic nonlinear state-space models, in this paper we focus on a state-space model commonly used for neural spike rasters.

Consider an experiment with $R$ successive trials, during which we record the activity of $N$ neuronal spiking units. For each trial, let $(0, \mathcal{T}]$ be the continuous observation interval following the delivery of an exogenous stimulus at time $\tau = 0$. For each trial $r = 1, \ldots, R$ and neuron $n = 1, \ldots, N$, let $\idx[r] S n$ be the total number of spikes from neuron $n$ during trial $r$, and the sequence $0 < \tau^{(n)}_{r,1}<\ldots < \tau^{(n)}_{r,S_r^{(n)}}$ correspond to the times at which events from the neuronal unit occur. We assume that {{$\{\tau^{(n)}_{r,s}\}_{s=1}^{S_r^{(n)}}$}} is the realization in $(0, \mathcal{T}] $ of a stochastic point-process with counting process $N_r^{(n)}(\tau) = \int_{0}^\tau dN_r^{(n)}(u)$, 
where $dN_r^{(n)}(\tau)$ is the indicator function in $(0, \mathcal{T}]$ of {{$\{\tau^{(n)}_{r,s}\}_{s=1}^{S_r^{(n)}}$}}. A point-process is fully characterized by its conditional intensity function (CIF)~\citep{vere2003introduction}. Assuming all trials are i.i.d. realizations of the same point-process, the CIF $\lambda^{(n)}(\tau \given H_\tau)$ of $dN_r^{(n)}(\tau)$ for $r = 1, \ldots, R$ is
\begin{align}
	&\lambda^{(n)}(\tau \given H_\tau^{(n)})  \\ \nonumber
	&=\lim_{\delta \rightarrow 0} \frac{p\left(N_r^{(n)}(\tau+\delta)-N_r^{(n)}(\tau)=1 \given H_\tau^{(n)}\right)}{\delta},
\end{align}
where $H_\tau^{(n)}$ is the event history of the point process up to time $\tau$ for neuron $n$.   Suppose we sample $dN_r^{(n)}(\tau)$ at a resolution of $\delta$ to yield a binary event sequence.  We denote by $\{\Delta N_{t,r}^{(n)}\}_{t=1,r=1}^{T,R}$ the discrete-time process obtained by counting the number of events in $T =  \left\lfloor \mathcal{T} / \Delta \right \rfloor$ disjoint bins of width $\Delta = M \cdot \delta$, where $M \in \mathbb{N}$.  Given $x_0^{(n)}$ and  $\idx[0] \psi n$, a popular approach is to encode a discrete-time representation of the CIF $\{\idx[t] \lambda n\}_{t=1}^T$ within an autoregressive process that underlies a binomial state-space model with observations {{$\{\Delta N_{t,r}^{(n)}\}_{t=1,r=1}^{T,R}$}} \citep{smith2003estimating}: 
\begin{align}
	 &x_1^{(n)} \given \idx \tmu n \sim \mathcal{N}(x_0^{(n)} + \idx \tmu n, \idx[0] \psi n), \label{bssm} \\ 
	  &x_t^{(n)} \given x_{t-1}^{(n)}, \tilde{\psi}^{(n)} \sim \mathcal{N} (x_{t-1}^{(n)}, \tpsi^{(n)}),  &1 < t, \nonumber \\ 
       &\idx[t] p n = \idx[t] \lambda n \cdot \delta = \sigma( \idx[t] x n) = \frac{\exp {x_t^{(n)}}}{1+ \exp {x_t^{(n)}}}, &1 \leq t, \nonumber \\
     & y_t^{(n)}  = \sum\limits_{r=1}^R  \Delta N_{t,r}^{(n)} \sim \text{Bin}\left(R \cdot M, \idx[t] p n\right) , &1 \leq t \nonumber.
\end{align}
where $\bidx \ttheta n = [\idx \tmu n, \log \idx \tpsi n]^\top$ are the parameters of interest.  We can cluster the neuronal units by these parameters by assuming that they arise from the Dirichlet process mixture of Equation \ref{crp}, in which $\bidx \theta k = [\idx \mu k, \log \idx \psi k]^\top = \bidx \ttheta n$ for all $n$ such that $\idx z n = k$.  

The parameter $\mu^{(k)} $ describes the extent to which the exogenous stimulus modulates the response of the neuron -- a positive value of $\mu^{(k)}$ indicates excitation, a negative value indicates inhibition, and a value close to zero indicates no response. The state transition density $f$ imposes a stochastic smoothness constraint on the CIF of neuron $n$, where $\psi^{(k)}$ controls the degree of smoothness.  A small value of $\psi^{(k)}$ suggests that the neurons exhibit a sustained change in response to the stimulus, whereas a large value of $\psi^{(k)}$ indicates that the change is unsustained.  With respect to the DPnSSM, the goal is to cluster the neurons according to the extent of the initial response $\mu^{(k)}$ and how sustained the response is $\psi^{(k)}$ . 

\section{INFERENCE ALGORITHM}
For conducting posterior inference on the DPnSSM, we introduce a Metropolis-within-Gibbs sampling procedure inspired by Algorithm 8 from \cite{neal2000markov}.  We derive the following process for alternately sampling (1) the cluster assignments $Z \given \bd \Theta, \bd Y, \alpha, G$ and (2) the cluster parameters $\bd \Theta \given Z, \bd Y, \alpha, G$.  A summary of the inference algorithm is given in Algorithm \ref{alg}. Outputs are samples $(\aidx Z i, \aidx {\bd \Theta} i)$ for iterations $i = 1, 2, \ldots, I$.

For any set $\mathcal{S} = \{\idx s 1, \ldots, \idx s J\}$, we denote set $\mathcal{S}$ without the $j$-th element as $\idx {\mathcal{S}} {-j} = \mathcal{S} \setminus \{\idx s j\}$.

\subsection{Sampling Cluster Assignments}
For a given time series $n \in \{1, \ldots, N\}$, we sample its cluster assignment from the distribution: 
\begin{align}
p(\idx z n &\given \idx Z {-n}, \bd \Theta, \bd Y, \alpha, G) \label{clust-assign} \\ 
&\propto p(\idx z n \given \idx Z {-n}, \bd \Theta, \alpha, G) \cdot p(\bd Y \given Z, \bd \Theta, \alpha, G) \nonumber \\
&\propto p(\idx z n \given \idx Z {-n}, \alpha) \cdot p(\bidx y n \given \idx z n, \bd \Theta, G). \nonumber
\end{align}
The first term in Equation \ref{clust-assign} can be represented by the categorical distribution,
\begin{numcases}{p(\idx z n = k) = }
\dfrac{\idx N {k}}{N - 1 + \alpha},  \hspace{0.3em} k = 1, \ldots, K',  \label{discrete}\\
\dfrac{\alpha / m}{N - 1 + \alpha}, \hspace{0.3em} k = K' + 1, \ldots, K' + m, \nonumber
\end{numcases}
where $K'$ is the number of unique $k$ in $\idx Z {-n}$, $\idx N {k}$ is the number of cluster assignments equal to $k$ in $\idx Z {-n}$, and $m \geq 1$ is some {integer algorithmic parameter.} For brevity, we drop the conditioning on $\idx Z {-n}$ and $\alpha$.

The second term in Equation \ref{clust-assign} is equivalent to the parameter likelihood $p(\bidx y n \given \bidx \theta k)$, where $\bidx \theta k$ is known if $k \in \{1, \ldots, K'\}$; otherwise, $\bidx \theta k$ must first be sampled from $G$ if $k \in \{K'+1, \ldots K'+m\}$.  Since $\bidx y n$ is the output of a nonlinear state-space model, we must use particle methods to approximate this parameter likelihood. We employ a recently proposed method known as controlled sequential Monte Carlo (cSMC) to produce low-variance estimates of this likelihood for a fixed computational cost~\citep{heng2017controlled}.  We outline the basic premise behind cSMC in Section \ref{ssec:csmc}.


\subsection{Sampling Cluster Parameters}
For a given cluster $k \in \{1, \ldots, K\}$, we wish to sample from the distribution:
\begin{align}
p(\bidx \theta k &\given \bidx \Theta {-k}, Z, \bd Y, \alpha, G) \label{clust-param} \\
&\propto p(\bidx \theta k \given \bidx \Theta {-k}, Z, \alpha, G) \cdot p(\bd Y \given \bd \Theta, Z, \alpha, G) \nonumber \\
&\propto p(\bidx \theta k  \given G) \cdot \prod_{n \given \idx z n = k} p(\bidx y n \given \bidx \theta k). \nonumber
\end{align}
The first term of Equation \ref{clust-param} is the probability density function of the base distribution, and the second term is a product of parameter likelihoods. Because the likelihood conditioned on class membership involves integration of the state sequence $\bidx x n$, and the prior $G$ is on the parameters of the state sequence, marginalization destroys any conjugacy that might have existed between the state sequence prior and parameter priors. 

To sample from the conditional posterior of parameters given cluster assignments, \cite{middleton2014} re-introduces the state sequence as part of his sampling algorithm for the linear Gaussian state-space case. 
We use an approach that obviates the need to re-introduce the state sequence and generalizes to scenarios where the prior on parameter and the state sequence may not have any conjugacy relationships. In particular, our sampler uses a Metropolis-Hastings step with proposal distribution $r(\bd \theta' \given \bd \theta)$ to sample from the class conditional distribution of parameters given cluster assignments. This effectively becomes one iteration of the well-known particle marginal Metropolis-Hastings (PMMH) algorithm \citep{andrieu2010particle}.  To evaluate the second term of Equation \ref{clust-param} for PMMH, we once again choose to use cSMC (Section \ref{ssec:csmc}).   

\subsection{Controlled Sequential Monte Carlo} \label{ssec:csmc}
Controlled SMC is based on the idea that we can modify a state-space model in such a way that standard bootstrap particle filters \citep{doucet2001introduction} give lower variance estimates while the likelihood of interest is kept unchanged. More precisely, the algorithm introduces a collection of positive and bounded functions $\gamma = \{\gamma_1, \ldots, \gamma_T\}$, termed a policy, that alter the transition densities of the model in the following way:
\begin{align}
h^\gamma(x_1; \bd \theta) &\propto h(x_1; \bd \theta) \cdot \gamma_1(x_1), \\
f^\gamma_t(x_{t-1}, x_{t}; \bd \theta) &\propto f(x_{t-1}, x_t; \bd \theta) \cdot \gamma_t(x_t), & 1 < t. \nonumber
\end{align}
To ensure that the likelihood associated with the modified model is the same as the original one, we introduce a modified version of the state-dependent likelihood $g$, denoted by $g_{1}^{\gamma},\ldots,g_{T}^{\gamma}$. On the modified model defined by $h^{\gamma}, \{f_{t}^{\gamma}\}_{t=2}^T,\{g_{t}^{\gamma}\}_{t=1}^T$, we can run a bootstrap particle filter and compute the likelihood estimator:
\begin{align}
\hat{p}^{\gamma}(\bd y \given \bd \theta) = \prod_{t=1}^\text{T} \left(\frac{1}{S} \sum_{s=1}^S g_t^\gamma(x_t^s, y_t; \bd \theta)\right),
\end{align}
where $S$ is the number of particles and $x_t^s$ is the $s$-th particle at time $t$.  The policy $\gamma$ can be chosen so as to minimize the variance of the above likelihood estimator; the optimal policy minimizing that variance is denoted by $\gamma^{*}$.

When $h, f$ are Gaussian and $g$ is log-concave with respect to $x_t$ (such as in the point-process state-space model of Equation \ref{bssm}), we can justify the approximation of $\gamma^*$ with a series of Gaussian functions. This allows us to solve for $h^{\gamma}, \{f_{t}^{\gamma}\}_{t=2}^T,\{g_{t}^{\gamma}\}_{t=1}^T$ using an approximate backward recursion method that simply reduces to a sequence of constrained linear regressions. We provide a more rigorous treatment of the exact details in \textbf{Appendix \ref{csmc-app}}.  

Starting from an initial policy {$\aidx \gamma 0$}, we can thus run a first bootstrap particle filter and obtain an approximation {$\aidx \gamma 1$} of $\gamma^{*}$. One can then iterate $L$ times to obtain refined policies, and consequently, lower variance estimators of the likelihood. Our empirical testing demonstrates that cSMC can significantly outperform the standard BPF in both precision and efficiency, while keeping $L$ very small. This justifies its use in the DPnSSM inference algorithm.

\begin{algorithm}
  \begin{algorithmic}[1]
    \For{$i = 1, \ldots, I$}
    	\State{Let $Z = \aidx Z {i-1}$ and $\bd \Theta = \aidx {\bd \Theta} {i-1}$.}
	\item[\quad \quad \emph{// Sample cluster assignments.}]
    	\For{$n = 1, \ldots, N$}
		\State{Let $K'$ be the number of distinct $k$ in $\idx Z {-n}$.}
		\For{$k = 1, \ldots, K' + m$}
			\State{Run cSMC to compute $p(\bidx y n \given \bidx \theta k)$.}
		\EndFor
		\State{Sample $\idx z n \given \idx Z {-n}, \bd \Theta, \bd Y, \alpha, G$.}
	\EndFor
	\State{Let $K$ be the number of distinct $k$ in $Z$.}
	\item[\quad \quad \emph{// Sample cluster parameters.}]
	\For{$k = 1, \ldots, K$}
		\State{Sample proposal $\bd \theta' \sim r(\bd \theta' \given \bidx \theta k)$.}
		\For{$n \in \{1, \ldots, N\} \given \idx z n = k$}
			\State{Run cSMC to compute $p\left(\bidx y n \given \bd \theta' \right)$.}
		\EndFor
		\State{Let $a = \frac{p(\bd \theta' \given \bidx \Theta {-k}, Z, \bd Y, \alpha, G) \cdot r(\bidx \theta k \given \bd \theta ')}{p(\bidx \theta k \given \bidx \Theta {-k}, Z, \bd Y, \alpha, G) \cdot r(\bd \theta' \given \bidx \theta k)}$.}
		\State{Let $\bidx \theta k = \bd \theta'$ with probability $\min(a, 1)$.}
	\EndFor
	\State{Let $\aidx Z i = Z$ and $\aidx {\bd \Theta} i = \bd \Theta$.}
  \EndFor \\
  \Return $(\aidx Z 1, \baidx \Theta 1), \ldots, (\aidx Z I, \baidx \Theta I)$
  \end{algorithmic}
  \caption{\texttt{InferDPnSSM}($\bd Y, \alpha, G, m, r, I, \aidx Z 0, \baidx \Theta 0$)} \label{alg}
\end{algorithm}

\section{RESULTS}
We investigate the ability of the DPnSSM to cluster time series from simulated and real neuronal rasters.\footnote{Python code for all experiments can be found at \texttt{https://github.com/ds2p/state-space-mixture}.}

\subsection{Selecting Clusters} \label{ssec:sel-clust}
The output of Algorithm 1 is a set of Gibbs samples $(\aidx Z 1, \baidx \Theta 1), \ldots, (\aidx Z I, \baidx \Theta I)$.  Each sample $(\aidx Z i, \baidx \Theta i)$ may very well use a different number of clusters.  The natural question that remains is how to select a single final clustering $(Z^*, \bd \Theta^*)$ of our data from this output.  There is a great deal of literature on answering this subjective question.  We follow the work of \cite{dahl2006model} and \cite{nieto2014bayesian}.  

Each Gibbs sample describes a clustering of the time series; we therefore frame the objective as selecting the most representative sample from our output.  To start, we take each Gibbs sample $i$ and construct an $N \times N$ co-occurrence matrix $\baidx \Omega i$ in which, 
\begin{align}
\baidx [(n, n')] \Omega i =
\begin{cases}
1, & \idx z n = \idx z {n'} \given \idx z n, \idx z {n'} \in \aidx Z i,\\
0, & \idx z n \neq \idx z {n'} \given \idx z n, \idx z {n'} \in \aidx Z i.
\end{cases} \label{co-occurrence}
\end{align}
This is simply a matrix in which the $(n, n')$ entry is 1 if series $n$ and series $n'$ are in the same cluster for the $i$-th Gibbs sample and 0 otherwise.  {We then define {$\bd \Omega = (I-B)^{-1} \sum_{i=B+1}^I \baidx \Omega i$}} as the mean co-occurrence matrix, where $B \geq 1$ is the number of pre-burn-in samples.  This matrix summarizes information from the entire trace of Gibbs samples.  The sample $i^*$ that we ultimately select is the one that minimizes the Frobenius distance to this matrix, i.e. $i^* = \arg \min_{i} \norm{\baidx \Omega i - \bd \Omega}_F$.
We use the corresponding assignments and parameters as the final  clustering $(Z^*, \bd \Theta^*) = (\aidx Z {i^*}, \baidx \Theta {i^*})$.  The appeal of this procedure is that it makes use of global information from all the Gibbs samples, yet ultimately selects a single clustering produced by the model.  If there are multiple Gibbs samples $i_1, \ldots, i_J$ such that $\baidx \Omega {i^*} = \baidx \Omega {i_1} = \ldots = \baidx \Omega {i_J}$, then we redefine $\bd \Theta^*$ as a simple average, as explained in \textbf{Appendix \ref{mult-co}}.     

\subsection{Simulated Neural Spiking Data} \label{ssec:sim-data}
We conduct a simulated experiment to test the ability of the DPnSSM to yield desired results in a setting in which the ground truth clustering is known. 

\subsubsection{Data Generation} \label{sssec:sim-data}
We simulate $N = 25$ neuronal rasters that each record data for $R = 45$ trials over the time interval $(-500, \mathcal{T}]$ milliseconds (ms) before/after an exogenous stimulus is applied at 0 ms, where $\mathcal{T} = 1500$.  For each trial, the resolution of the binary event sequence is $\delta = 1$ ms.  We create bins of size  $\Delta =  M \cdot \delta$, where $M = 5$, and observe neuron $n$ firing $\Delta \idx[t, r] N n \leq M$ times during the $t$-th discrete time interval $(t\Delta - \Delta, t\Delta]$ for trial $r$.  

We use the following process for generating the simulated data:  For each neuron $n$, the initial rate is independently drawn as $\idx \lambda n \sim \text{Uniform}(10, 15)$ Hz.  Each neuron's \emph{type} is determined by the evolution of its discrete-time CIF  $\idx[t] \lambda n $ over time.  We split the discrete-time intervals into three parts -- $\bd t_1 = \{-99, \ldots, 0\}$, $\bd t_2 = \{1, \ldots, 50\}$, and $\bd t_3 = \{51, \ldots, 300\}$. We generate five neurons from each of the following five types: 
\begin{enumerate}[noitemsep,nolistsep]
\item \emph{Excited, sustained} neurons with rate $\idx[t] \lambda n = \idx \lambda n$ for $t \in \bd t_1$; rate $\idx[t] \lambda n = \idx \lambda n \cdot \exp(1)$ for $t \in \bd t_2, \bd t_3$.
\item \emph{Inhibited, sustained} neurons with rate $\idx[t] \lambda n = \idx \lambda n$ for $t \in \bd t_1$; rate $\idx[t] \lambda n = \idx \lambda n \cdot \exp(-1)$ for $t \in \bd t_2, \bd t_3$.
\item \emph{Non-responsive} neurons with rate $\idx[t] \lambda n = \idx \lambda n$ for $t \in \bd t_1, \bd t_2, \bd t_3$.
\item \emph{Excited, unsustained} neurons with rate $\idx[t] \lambda n = \idx \lambda n$ for $t \in \bd t_1$; rate $\idx[t] \lambda n = \idx \lambda n \cdot \exp(1)$ for $t \in \bd t_2$; rate $\idx[t] \lambda n = \idx \lambda n$ for $t \in \bd t_3$.
\item \emph{Inhibited, unsustained} neurons with rate $\idx[t] \lambda n = \idx \lambda n$ for $t \in \bd t_1$; rate $\idx[t] \lambda n = \idx \lambda n \cdot \exp(-1)$ for $t \in \bd t_2$; rate $\idx[t] \lambda n = \idx \lambda n$ for $t \in \bd t_3$.
\end{enumerate}
Following the point-process state-space model of Equation \ref{bssm} -- which assumes i.i.d. trials -- we simulate, 
\begin{align}
\idx[t] y n = \sum_{r=1}^{R=45} \Delta \idx[t, r] N n &\sim \text{Bin}(R \cdot M = 225, \idx[t] p n),
\end{align}
where $\idx[t] p n = \idx[t] \lambda n \cdot \delta$ for $t = -99, \ldots, 300$.  These are the observations $\bd Y = \{\bidx y 1, \ldots, \bidx y N\}$ that are fed to the DPnSSM.  The model is then tasked with figuring out the original ground-truth clustering.

\subsubsection{Modeling} \label{sim-model-details}
In modeling these simulated data as coming from the DPnSSM, we employ the generative process specified by Equation \ref{bssm}; that is,
\begin{align} \label{sim-model}
\idx[1] x n \given \idx \mu k &\sim \Norm(\idx[0] x n + \idx \mu k, \psi_0),  \\ \nonumber
\idx[t] x n \given \idx[t-1] x n, \idx \psi k &\sim \Norm(\idx[t-1] x n, \idx \psi k), & & 1 < t \leq T, \\ \nonumber
\idx[t] y n \given \idx[t] x n &\sim \text{Bin}(225, \sigma(\idx[t] x n)), & & 1 \leq t \leq T, 
\end{align} 
where cluster parameters are $\bidx \theta k = [\idx \mu k, \log \idx \psi k]^\top$.  
 
The series are fed into Algorithm \ref{alg} with hyperparameter values $\alpha = 1$, $G[\mu, \psi] = [\Norm(0, 2), \text{Unif}(-15, 0)]$, and $m = 5$.  For every series $n$, we compute the initial state $\idx[0] x n = \sigma^{-1}(1 / 500 \cdot \sum_{t=-99}^{0} \idx[t] y n)$ from the observations before the stimulus in that series.  In addition, we let $\psi_0 = 10^{-10}$, a very small value that forces any change in the latent state at $t = 1$ to be explained by the cluster parameter $\idx \mu k$.  The initial clustering $(\aidx Z 0, \baidx \Theta 0) = (\bd 1, \bd \theta_0)$, where $\bd 1$ is a vector of $N$ ones denoting that every series begins in the same cluster and $\bd \theta_0$ is sampled from $G$.  For the proposal $r(\bd \theta' \given \bd \theta)$, we use a $\Norm(\bidx \theta k,  0.25 \cdot {\bold I})$ distribution, where $\bold I$ is the $2 \times 2$ identity matrix.  We run the sampling procedure for $I$ = 10,000 iterations and apply a burn-in of $B$ = 1,000 samples. To compute likelihood estimates, we use $L = 3$ cSMC iterations and $S = 64$ particles.  

\begin{figure}[h]
\begin{center}
\includegraphics[scale=0.3]{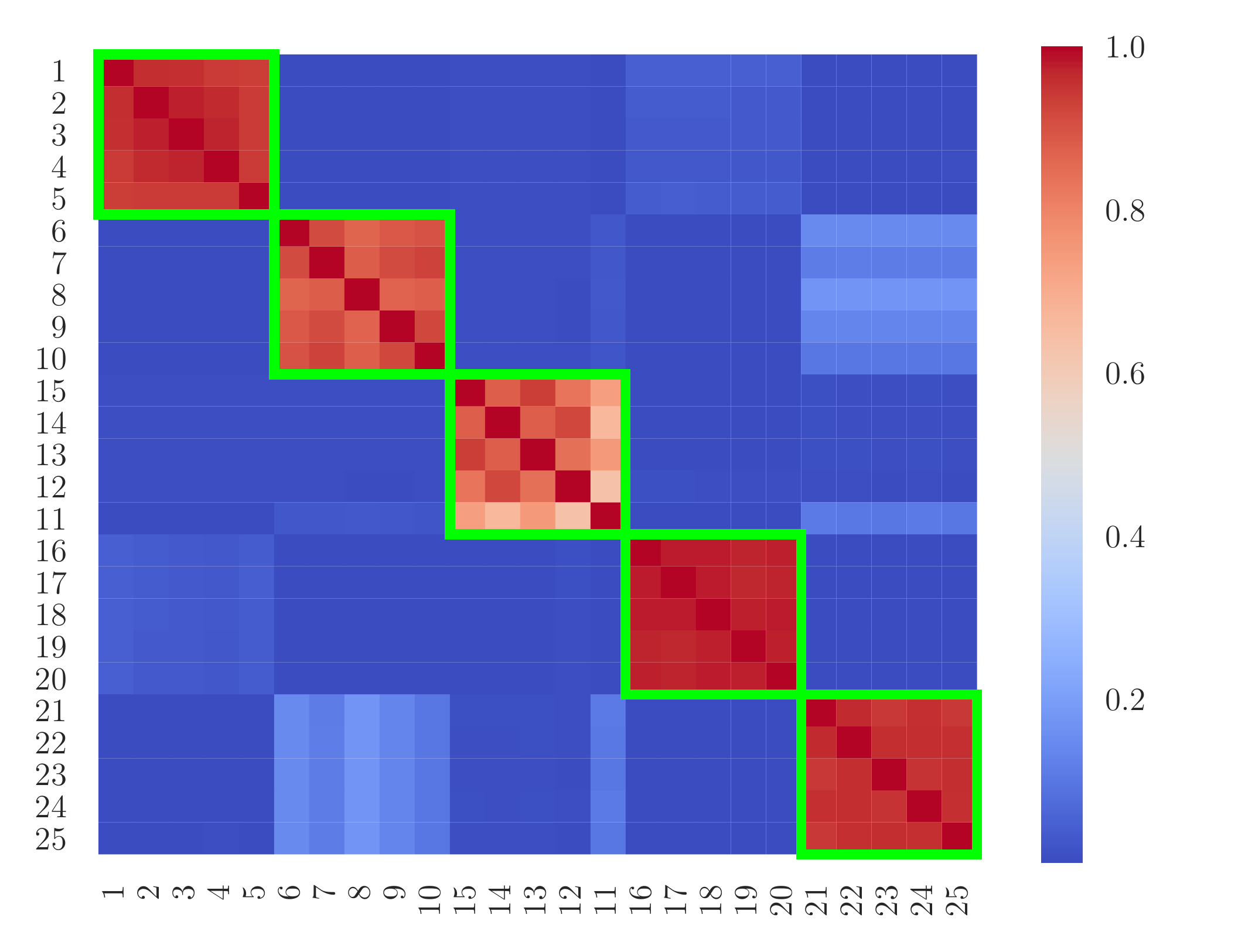}
\end{center}
\caption{Heatmap of mean co-occurrence matrix $\bd \Omega$ for simulation results.  Elements of the selected co-occurrence matrix $\baidx \Omega {i^*}$ that are equal to 1 are enclosed in green squares.  Each square corresponds to a distinct cluster.}
\label{sim-hm}
\end{figure}

A heatmap of the resultant mean co-occurrence matrix $\bd \Omega$ (Equation \ref{co-occurrence}) and the selected clustering $\baidx \Omega {i^*}$ can be found in Figure \ref{sim-hm}.  The rows and columns of this matrix have been reordered to aid visualization of clusters along the diagonal of $\bd \Omega$.  From this experiment, we can see that the DPnSSM inference algorithm is able to successfully recover the five ground-truth clusters.  \textbf{Appendix \ref{sim-robust}} present some results on the robustness of the model to misspecification of the stimulus onset.     

Table \ref{sim-table} summarizes the final cluster parameters $\bd \Theta^*$.  As one may expect, a highly positive $\idx {\mu^*} k$ corresponds to neurons that are excited by the stimulus, while a highly negative $\idx {\mu^*} k$ corresponds to neurons that are inhibited.  The one cluster with $\idx {\mu^*} k \approx 0$ corresponds to non-responsive neurons.  With $\idx {\mu^*} k$, the algorithm is able to approximately recover the true amount by which the stimulus increases or decreases the log of the firing rate/probability, which is stated in Section \ref{sssec:sim-data} -- i.e. $+1$ for $k \in \{1, 4\}$, $-1$ for $k \in \{2, 5\}$ and $0$ for $k = 3$.  This provides an interpretation of the numerical value of $\idx \mu k$.  Indeed, if $\exp \idx[0] x n < < 1, \exp \idx[1] x n << 1$, as is often the case when modeling neurons, then the expected increase in the log of the firing probability due to the stimulus is: 
\begin{align}
\E & \left[\log \frac{\sigma(\idx[1] x n)}{\sigma(\idx[0] x n)}\right] \approx \E \left[\log \frac{\exp \idx[1] x n}{\exp \idx[0] x n} \right] = \idx \mu k.
\end{align}   

In addition, the values for $\log \idx {\psi^*} k$ in Table \ref{sim-table} reveal that the algorithm uses this dimension to correctly separate unsustained clusters from sustained ones.  For $k \in \{1, 2, 3\}$, the algorithm infers smaller values of $\log \idx {\psi^*} k$ because the change in the firing rate is less variable after the stimulus has taken place, whereas the opposite is true for $k \in \{4, 5\}$.  In summary, the DPnSSM is able to recover some of the key properties of the data in an unsupervised fashion, thereby demonstrating its utility on this toy example.

\begin{table}[h]
\vspace{-3mm}
\caption{Parameters for simulation, where $\bidx {\theta^*}  k = [\idx {\mu^*} k, \log \idx {\psi^*} k]^\top$ for each $\bidx {\theta^*}  k \in \bd {\Theta^*}$.} \label{sim-table}
\begin{center}
\begin{tabular}{c|ccl}
$k$ & $\idx {\mu^*} k$ & $\log \idx {\psi^*} k$ &\textbf{Effect} \\
\hline
$1$ & $+0.96$ & $-10.88$ &  Excited, Sustained \\
$2$ &$-0.92$ & $-12.32$ & Inhibited, Sustained \\
$3$ &$+0.03$ & $-10.04$  & Non-responsive \\
$4$ &$+1.04$ & $-5.55$ & Excited, Unsustained\\
$5$ &$-0.89$ & $-5.89$ & Inhibited, Unsustained
\end{tabular}
\vspace{-4mm}
\end{center}
\end{table}
  
\subsection{Real Neural Spiking Data}
In addition to simulations, we produce clusterings on real-world neural spiking data collected in a fear-conditioning experiment designed to elucidate the nature of neural circuits that facilitate the associative learning of fear. The detailed experimental paradigm is described in~\cite{allsop2018corticoamygdala}.  In short, an observer mouse observes a demonstrator mouse receive conditioned cue-shock pairings through a perforated transparent divider. The experiment consists of $R = 45$ trials. During the first 15 trials of the experiment, both the observer and the demonstrator hear an auditory cue at time $\tau = 0$ ms. From trial 16 onwards, the auditory cue is followed by the delivery of a shock to the demonstrator at time $\tau$ = 10,000 ms, i.e. 10 seconds after the cue's administration. 

The data are recorded from various neurons in the prefrontal cortex of the observer mouse.  We apply our analysis to $N = 33$ neurons from this experiment that form a network hypothesized to be involved in the observational learning of fear.  Our time interval of focus is $(-500, \mathcal{T} = 1500]$ ms before/after the administration of the cue.  The raster data comes in the form of $\{\Delta \idx[t, r] N n\}_{t=1, r=1}^{T, R}$, binned at a resolution of $\Delta = 5$ ms with $T = 300$, where each $\Delta \idx[t, r] N n \leq M = 5$.  

We apply DPnSSM to identify various groups of responses in reaction to the auditory cue over time and over trials.  A group of neurons that respond significantly after trial 16 can be interpreted as one that allows the observer to understand when the demonstrator is in distress.

\subsubsection{Clustering Cue Data over Time} \label{sssec:time-data}
To cluster neurons by their cue responses over time, we collapse the raster for all neurons over the $R = 45$ trials.  Thus, for neuron $n$, define $\idx[t] y n = \sum_{r=1}^R \Delta \idx[t, r] N n$.  We apply the exact same model as the one used for the simulations (Equation \ref{sim-model}).  We also use all of the same hyperparameter values, as detailed in Section \ref{sim-model-details}.

\begin{figure}[h]
\begin{center}
\includegraphics[scale=0.3]{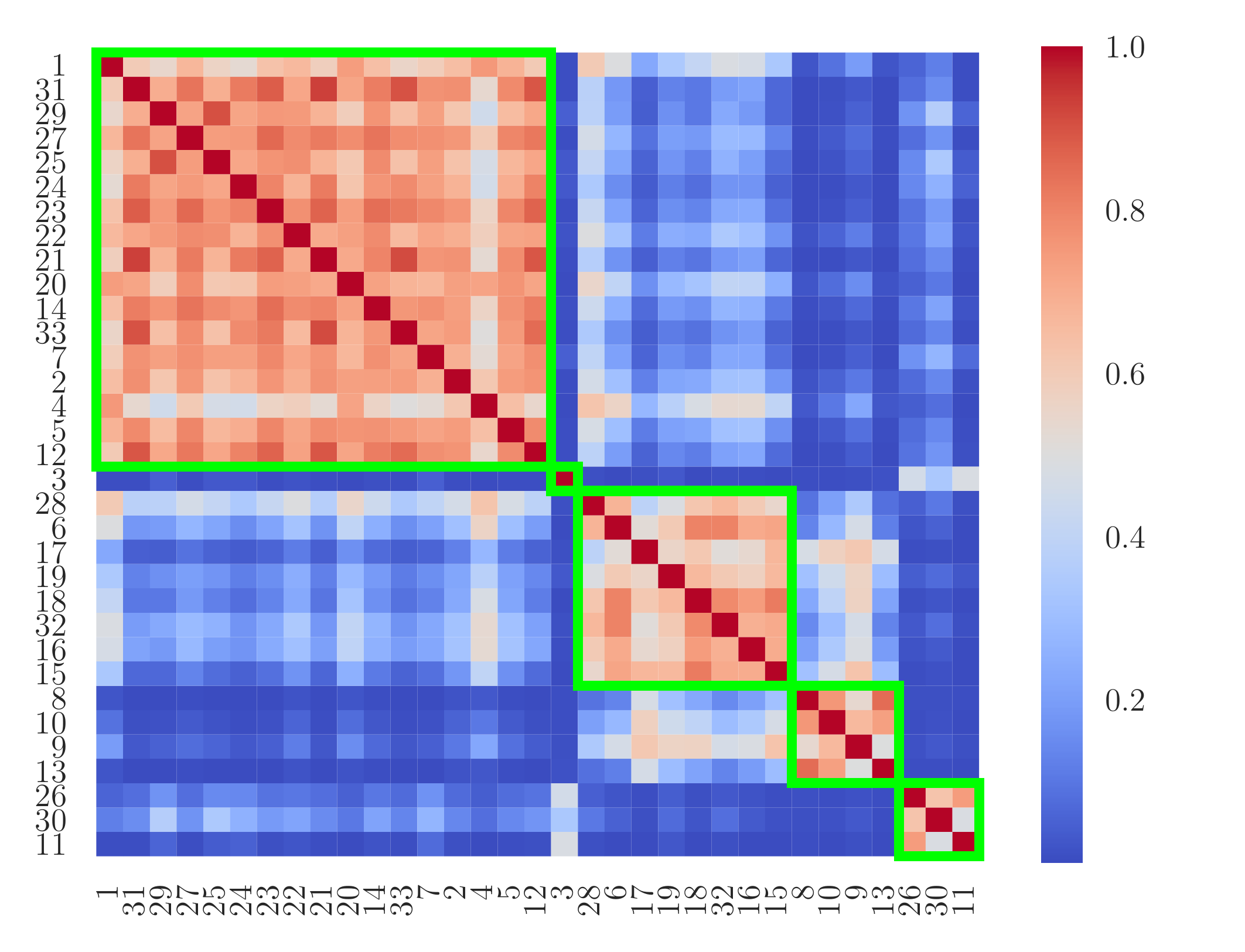}
\end{center}
\caption{Heatmap of mean co-occurrence matrix for cue data over time and selected clusters (green).} 
\label{cue-hm}
\end{figure}

\begin{figure}[h]
\begin{center}
\includegraphics[scale=0.45]{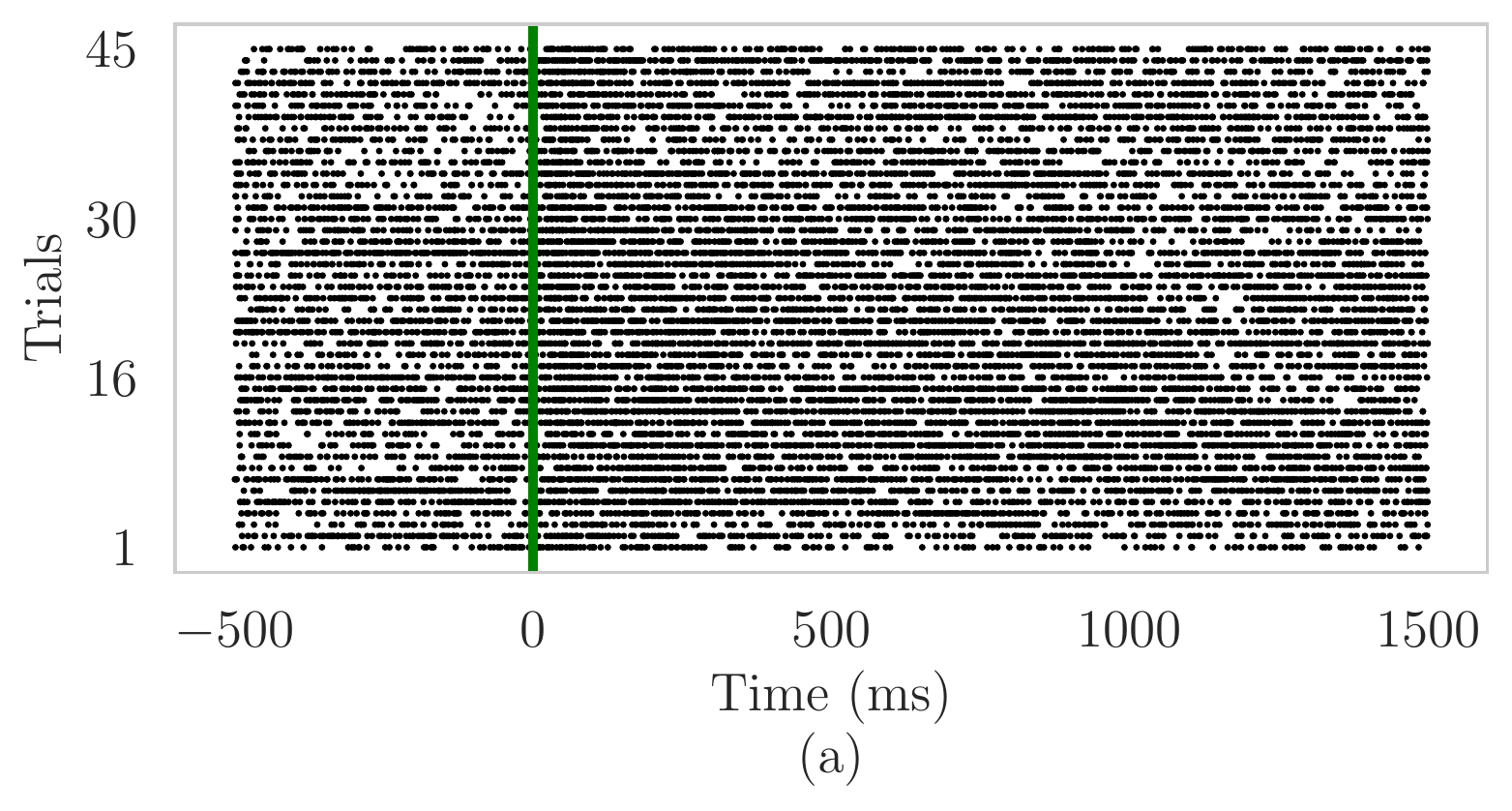}
\includegraphics[scale=0.45]{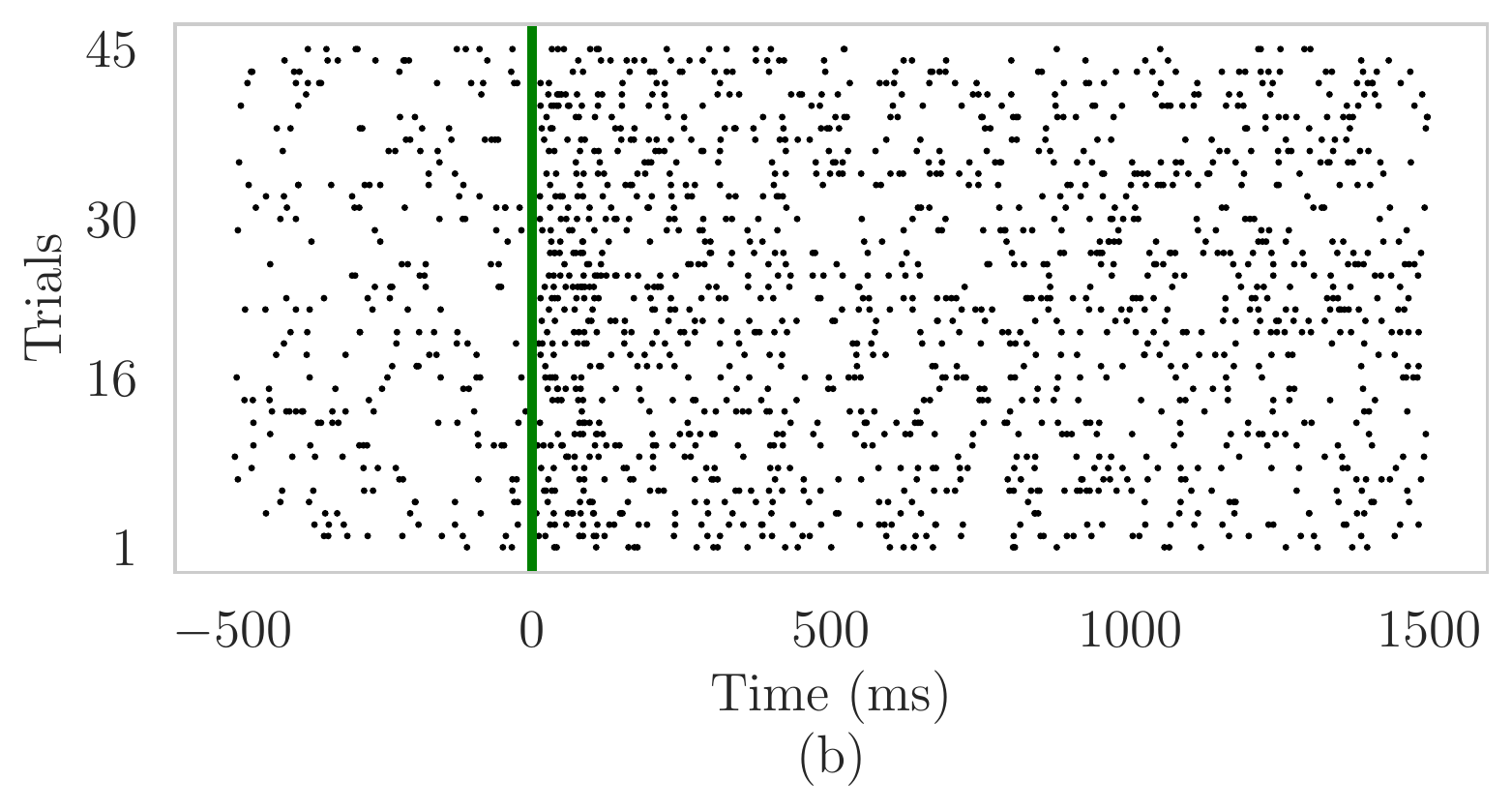}
\end{center}
\vspace{-4mm}
\caption{Overlaid raster plots of neuronal clusters with (a) moderately excited $(\idx {\mu^*} 1 = 0.54)$ and somewhat sustained ($\log \idx {\psi^*} 1 = -6.27$) responses to the cue; and (b) more excited $(\idx {\mu^*} 5 = 1.21)$ and less sustained responses ($\log \idx {\psi^*} 5 = -4.52$) to the cue.  A black dot at $(\tau, r)$ indicates a spike from one of the neurons in the corresponding cluster at time $\tau$ during trial $r$. The vertical green line indicates cue onset.} \label{cue-rasters}
\end{figure}

A heatmap of $\bd \Omega$ along with demarcations of $\baidx \Omega {i^*}$ for this experiment can be found in Figure \ref{cue-hm}. Overall, five clusters are selected by the algorithm.  Table \ref{cue-table} summarizes the chosen cluster parameters. Figure \ref{cue-rasters} shows two of the five clusters identified by the algorithm, namely those corresponding to $k = 1$ (Figure \ref{cue-rasters}a) and $k = 5$ (Figure \ref{cue-rasters}b). Figures for all other clusters can be found in \textbf{Appendix \ref{raster-plots}}.  Each of the figures was created by overlaying the rasters from neurons in the corresponding cluster. The fact that the overlaid rasters resemble the raster from a single unit (as opposed to random noise), with plausible parameter values in Table \ref{cue-table}, indicates that the algorithm has identified a sensible clustering of the neurons. 

The algorithm is able to successfully differentiate various types of responses to the cue as well as the variability of the responses. One advantage of not restricting the algorithm to a set number of classes a priori is that it can decide what number of classes best characterizes these data. In this case, the inference algorithm identifies five different clusters. We defer a scientific interpretation of this phenomenon to a later study. 

\begin{table}[h]
\vspace{-3mm}
\caption{Cluster parameters for cue data over time.} \label{sample-table} \label{cue-table}
\begin{center}
\begin{tabular}{c|cccl}
$k$ & $\idx {\mu^*} k$ & $\log \idx {\psi^*} k$ &\# of Neurons \\
\hline
$1$ & $+0.54$ & $-6.27$ & 17  \\
$2$ &$+0.03$ & $-6.80$ & 1 \\
$3$ &$-0.07$ & $-7.25$  & 8 \\
$4$ &$-0.70$ & $-6.01$ & 4 \\
$5$ &$+1.21$ & $-4.52$ & 3 
\end{tabular}
\vspace{-4mm}
\end{center}\end{table}   

\subsubsection{Clustering Cue Data over Trials} \label{cue-cluster}
We also apply DPnSSM to determine if neurons can be classified according to varying degrees of neuronal signal modulation when shock is delivered to another animal, as opposed to when there is no shock delivered. The shock is administered starting from the 16th trial onwards. Thus, to understand the varying levels of shock effect, we collapse the raster across the $T = 300$ time points (instead of the $R = 45$ trials, as was done in Section \ref{sssec:time-data}). In this setting, each $\idx[r] y n = \sum_{t=1}^T \Delta \idx[t, r] N n \in \{0, 1, \ldots, 2000\}$ represents the number of firings during the $r$-th trial.  For each neuron $n$, let the initial state be $\idx[0] x n = \sigma^{-1}(1 / 15 \cdot \sum_{r=1}^{15} \idx[r] y n)$.  Then, we use the following state-space model: 
\begin{align}
\idx[16] x n &\sim \Norm(\idx[0] x n + \idx \mu k, \psi_0),\\
\idx[r] x n \given \idx[r-1] x n &\sim \Norm(\idx[r-1] x n, \idx \psi k), & & 16 < r \leq R , \nonumber \\
\idx[r] y n \given \idx[r] x n &\sim \text{Bin}(2000, \sigma(\idx[r] x n)), & & 16 \leq r \leq R, \nonumber
\end{align} 
where once again the cluster parameters are $\bidx \theta k = [\idx \mu k, \log \idx \psi k]^\top$.  All other hyperparameter values are the same as those listed in Section \ref{sim-model-details}.  

\begin{figure}[h]
\begin{center}
\includegraphics[scale=0.3]{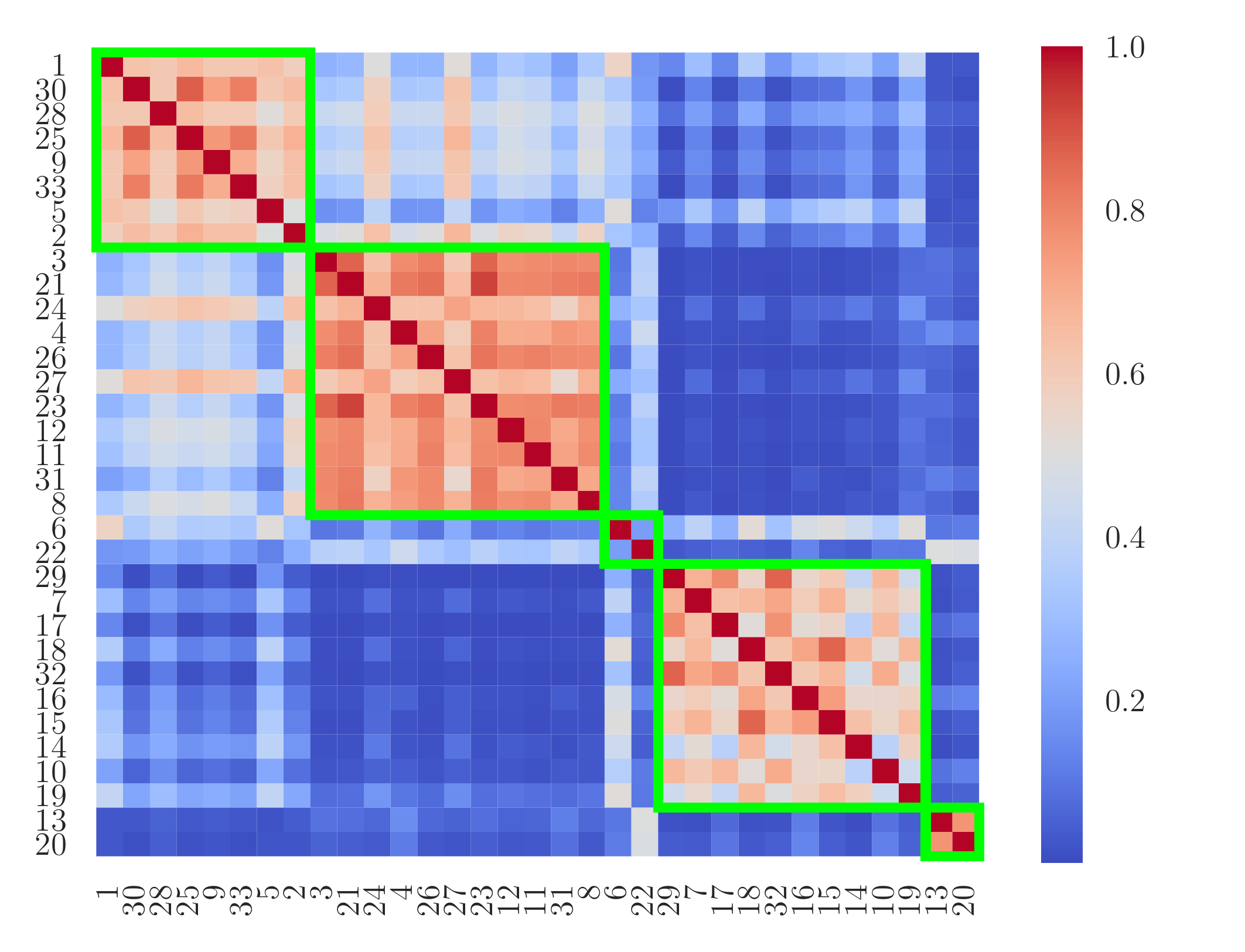}
\end{center}
\caption{Heatmap of mean co-occurrence matrix for cue data over trials and selected clusters (green).} 
\label{shock-hm}
\end{figure}

\begin{figure}[h]
\begin{center}
\includegraphics[scale=0.45]{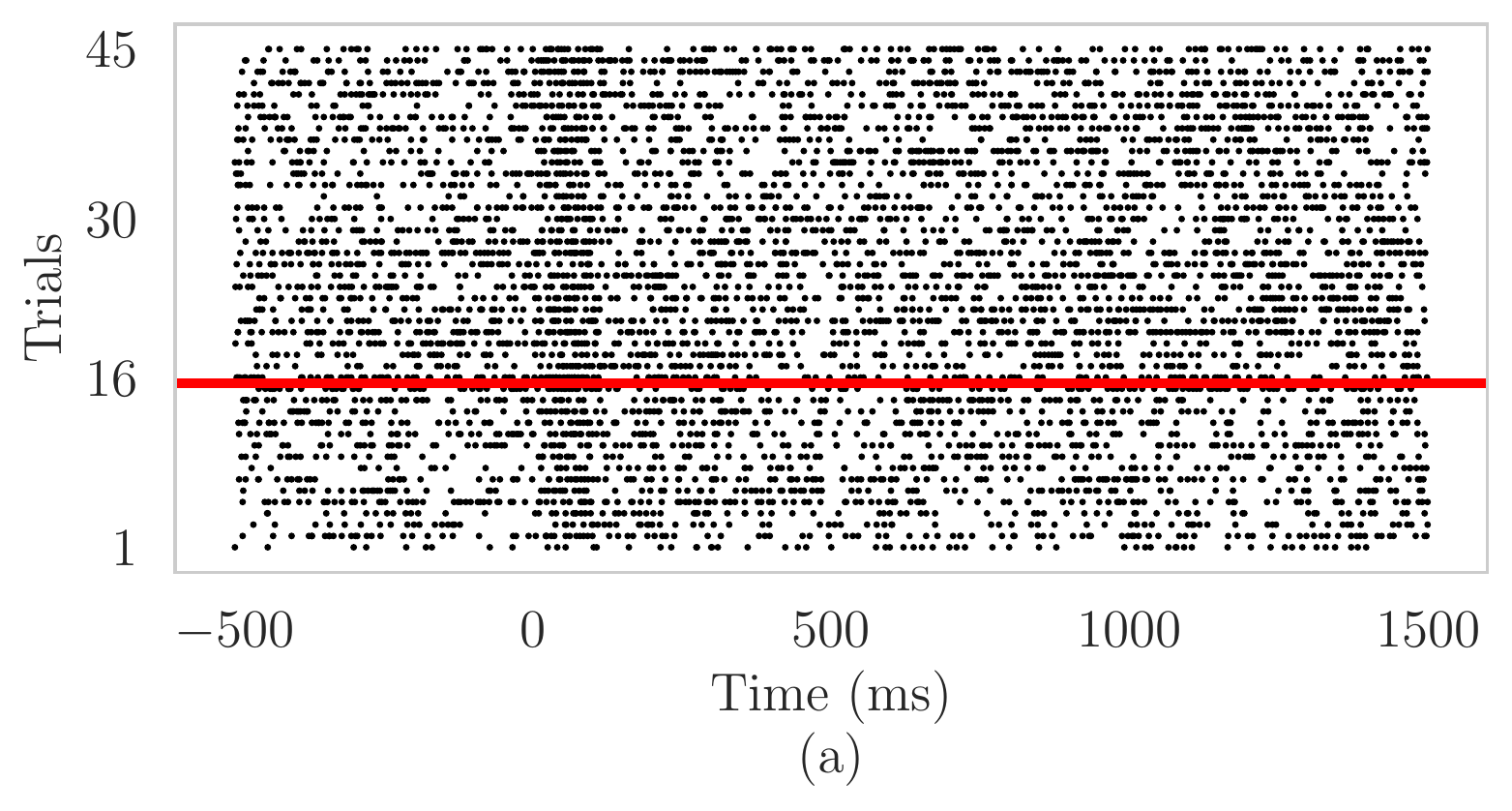}
\includegraphics[scale=0.45]{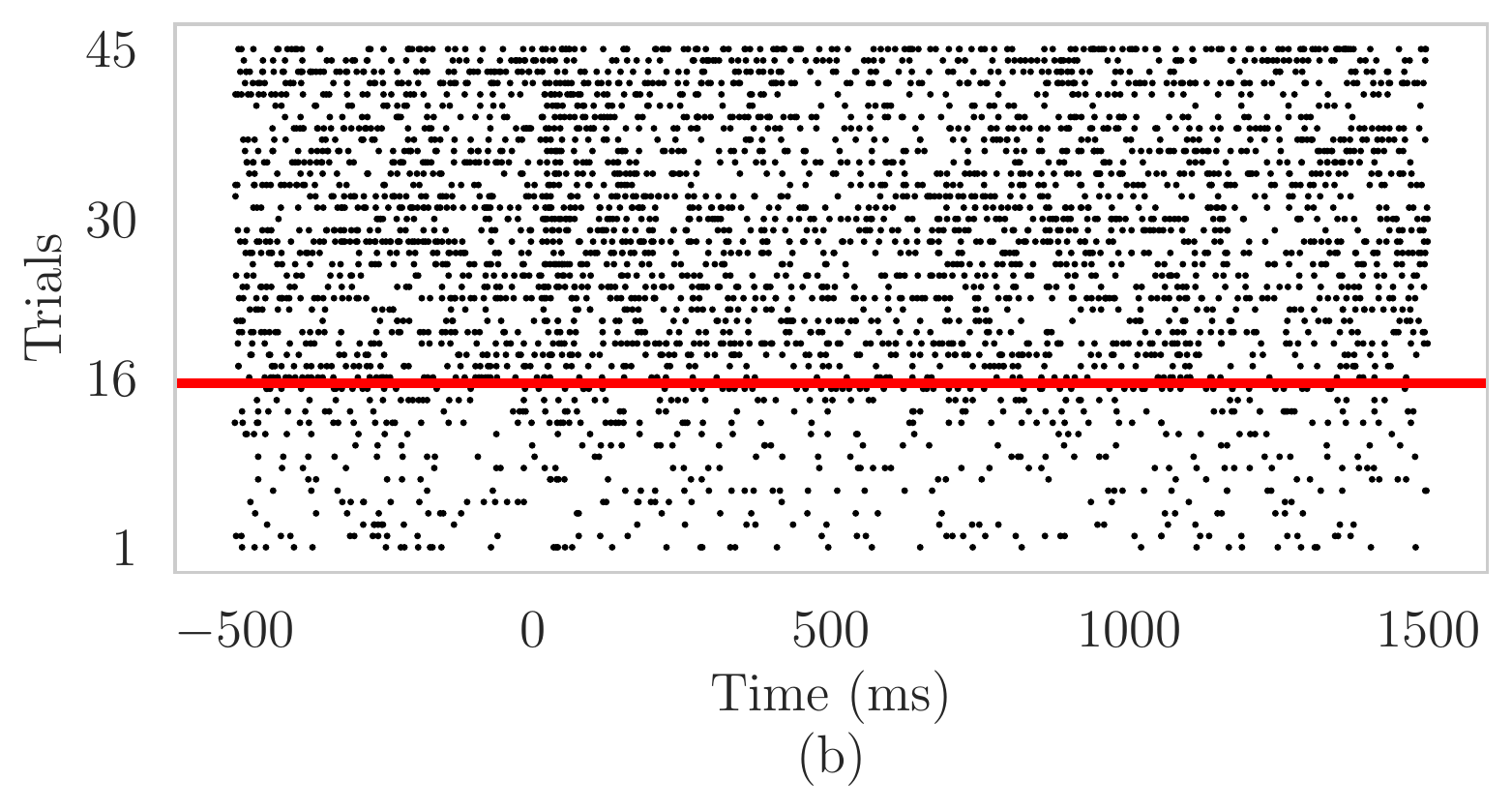}
\end{center}
\caption{(a) Overlaid raster plots of neuronal clusters with (a) slightly inhibited, variable responses ($k = 2)$ and (b) very excited, variable responses ($k = 4$). The red line marks the first trial with shock administration.} \label{shock-rasters}
\end{figure}

The corresponding heatmap, representative raster plots, and clustering results can be found in Figure \ref{shock-hm}, Figure \ref{shock-rasters}, and Table \ref{shock-table}, respectively. We speculate that the results suggest the existence of what we term \emph{empathy clusters}, namely groups of neurons that allow an observer to understand when the demonstrator is in distress. We will explore the implications of these findings to the neuroscience of observational learning of fear in future work.


\begin{table}[h]
\vspace{-3mm}
\caption{Cluster parameters for cue data over trials.} \label{shock-table}
\begin{center}
\begin{tabular}{c|cccl}
$k$ & $\idx {\mu^*} k$ & $\log \idx {\psi^*} k$ &\# of Neurons \\
\hline
$1$ & $+0.19$ & $-5.29$ & 8  \\
$2$ &$-0.14$ & $-4.40$ & 11 \\
$3$ &$-0.08$ & $-2.38$  & 2 \\
$4$ &$+0.87$ & $-2.95$ & 10 \\
$5$ &$-0.42$ & $-0.41$ & 2 
\end{tabular}
\vspace{-4mm}
\end{center}
\end{table} 

%

\subsection{Controlled SMC Versus BPF} \label{ssec:bpf-csmc}
Finally, we present some results on the advantages of using controlled sequential Monte Carlo over the bootstrap particle filter. Computing the parameter likelihood is a key task in Algorithm \ref{alg}. In each iteration, we perform $O(N \cdot K)$ particle filter computations during the sampling of the cluster assignments and another $O(N)$ particle filter computations during the sampling of the cluster parameters. Thus, for both the efficiency and precision of the algorithm, it is necessary to find a fast way to compute low-variance estimates. 

\begin{figure}[h!]
\begin{center}
\includegraphics[scale=0.42]{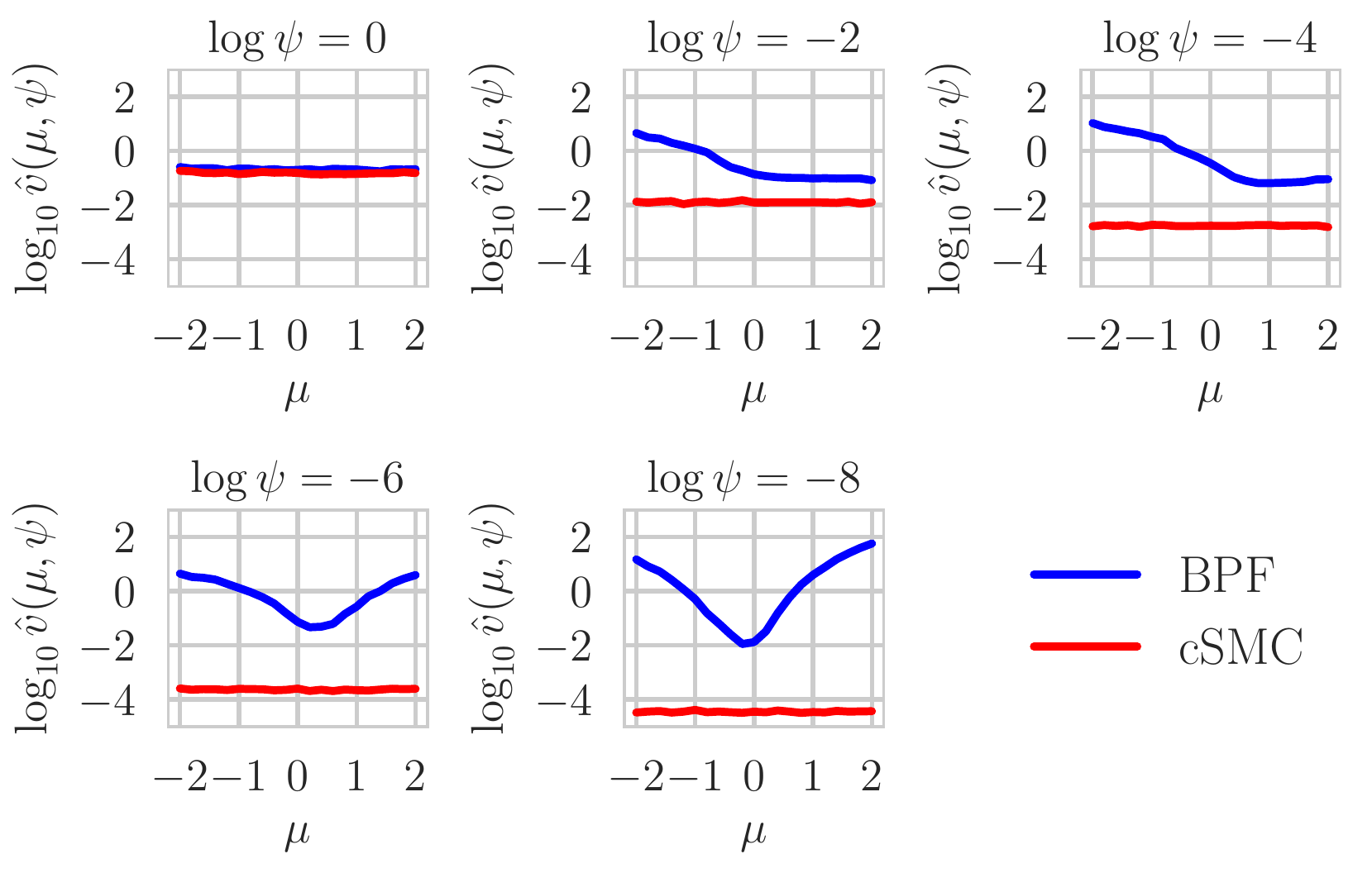}
\caption{Using BPF versus cSMC for parameter log-likelihood computation.  For each method, we plot an estimate of the variance $\hat{v}(\mu, \psi) \approx \Var[\log p(\bd y \given \bd \theta)]$, where $\bd \theta = [\mu, \log \psi]^\top$, over different values of $(\mu, \psi)$.} \label{bpf-csmc}
\end{center}
\end{figure}

Figure \ref{bpf-csmc} demonstrates the benefits of using cSMC over BPF for likelihood evaluation for a fixed computational cost.  Details on this experiment can be found in \textbf{Appendix \ref{csmc-bpf-app}}. In some cases, cSMC estimates have variances that are several orders of magnitude lower than those produced by BPF.  This is especially true for low values of the variability parameter $\log \psi$, which is crucial for this application since these are often the ones that maximize the parameter likelihood.  

\section{CONCLUSION}
We proposed a general framework to cluster time series with nonlinear dynamics modeled by nonlinear state-space models. To the best of the authors' knowledge, this is the first Bayesian framework for clustering time series that exhibit nonlinear dynamics. The backbone of the framework is the cSMC algorithm for low-variance evaluation of parameter likelihoods in nonlinear state-space models. We applied the framework to neural data in an experiment designed to elucidate the neural underpinnings of fear. We were able to identify potential clusters of neurons that allow an observer to understand when a demonstrator is in distress.

In future work, we plan to perform detailed analyses of the data from these experiments~\citep{allsop2018corticoamygdala}, and the implications of these analyses on the neuroscience of the observational learning of fear in mice. We will also explore applications of our model to data in other application domains such as sports and sleep research \citep{st2017modeling}, to name a few.

\newpage
\subsubsection*{Acknowledgements}
Demba Ba thanks Amazon Web Services (AWS), for their generous support and access to computational resources, and the Harvard Data Science Initiative for their support.  Pierre E. Jacob thanks the Harvard Data Science Initiative and the National Science Foundation (Grant DMS-1712872).  Kay M. Tye thanks the McKnight Foundation, NIH (Grant R01-MH102441-01), and NCCIH (Pioneer Award DP1-AT009925).

\bibliography{dpnssm}

\begin{thebibliography}{}

\bibitem[Allsop et~al., 2018]{allsop2018corticoamygdala}
Allsop, S.~A., Wichmann, R., Mills, F., Burgos-Robles, A., Chang, C.-J.,
  Felix-Ortiz, A.~C., Vienne, A., Beyeler, A., Izadmehr, E.~M., Glober, G.,
  et~al. (2018).
\newblock Corticoamygdala transfer of socially derived information gates
  observational learning.
\newblock {\em Cell}, 173(6):1329--1342.

\bibitem[Andrieu et~al., 2010]{andrieu2010particle}
Andrieu, C., Doucet, A., and Holenstein, R. (2010).
\newblock Particle {M}arkov chain {M}onte {C}arlo methods.
\newblock {\em Journal of the Royal Statistical Society: Series B (Statistical
  Methodology)}, 72(3):269--342.

\bibitem[Bauwens and Rombouts, 2007]{bauwens2007bayesian}
Bauwens, L. and Rombouts, J. (2007).
\newblock Bayesian clustering of many {GARCH} models.
\newblock {\em Econometric Reviews}, 26(2-4):365--386.

\bibitem[Brown et~al., 2004]{brown2004multiple}
Brown, E.~N., Kass, R.~E., and Mitra, P.~P. (2004).
\newblock Multiple neural spike train data analysis: state-of-the-art and
  future challenges.
\newblock {\em Nature Neuroscience}, 7(5):456.

\bibitem[Chiappa and Barber, 2007]{chiappa2007output}
Chiappa, S. and Barber, D. (2007).
\newblock Output grouping using {D}irichlet mixtures of linear {G}aussian
  state-space models.
\newblock In {\em Image and Signal Processing and Analysis, 2007. ISPA 2007.
  5th International Symposium on}, pages 446--451. IEEE.

\bibitem[Dahl, 2006]{dahl2006model}
Dahl, D.~B. (2006).
\newblock Model-based clustering for expression data via a {D}irichlet process
  mixture model.
\newblock {\em Bayesian Inference for Gene Expression and Proteomics}, 201:218.

\bibitem[Doucet et~al., 2001]{doucet2001introduction}
Doucet, A., De~Freitas, N., and Gordon, N. (2001).
\newblock An introduction to sequential {M}onte {C}arlo methods.
\newblock In {\em Sequential Monte Carlo Methods in Practice}, pages 3--14.
  Springer.

\bibitem[Durbin and Koopman, 2012]{durbin2012time}
Durbin, J. and Koopman, S.~J. (2012).
\newblock {\em Time Series Analysis by State Space Methods}, volume~38.
\newblock Oxford University Press.

\bibitem[Ferguson, 1973]{ferguson1973bayesian}
Ferguson, T.~S. (1973).
\newblock A {B}ayesian analysis of some nonparametric problems.
\newblock {\em The Annals of Statistics}, 1(2):209--230.

\bibitem[Guarniero et~al., 2017]{guarniero2017iterated}
Guarniero, P., Johansen, A.~M., and Lee, A. (2017).
\newblock The iterated auxiliary particle filter.
\newblock {\em Journal of the American Statistical Association},
  112(520):1636--1647.

\bibitem[Heng et~al., 2017]{heng2017controlled}
Heng, J., Bishop, A.~N., Deligiannidis, G., and Doucet, A. (2017).
\newblock Controlled sequential {M}onte {C}arlo.
\newblock {\em arXiv preprint arXiv:1708.08396}.

\bibitem[Humphries, 2011]{humphries2011spike}
Humphries, M.~D. (2011).
\newblock Spike-train communities: finding groups of similar spike trains.
\newblock {\em Journal of Neuroscience}, 31(6):2321--2336.

\bibitem[Inoue et~al., 2006]{inoue2006cluster}
Inoue, L.~Y., Neira, M., Nelson, C., Gleave, M., and Etzioni, R. (2006).
\newblock Cluster-based network model for time-course gene expression data.
\newblock {\em Biostatistics}, 8(3):507--525.

\bibitem[Middleton, 2014]{middleton2014}
Middleton, L. (2014).
\newblock Clustering time series: a {D}irichlet process mixture of
  linear-{G}aussian state-space models.
\newblock Master's thesis, Oxford University, United Kingdom.

\bibitem[Neal, 2000]{neal2000markov}
Neal, R.~M. (2000).
\newblock Markov chain sampling methods for {D}irichlet process mixture models.
\newblock {\em Journal of Computational and Graphical Statistics},
  9(2):249--265.

\bibitem[Nieto-Barajas and Contreras-Crist{\'a}n, 2014]{nieto2014bayesian}
Nieto-Barajas, L.~E. and Contreras-Crist{\'a}n, A. (2014).
\newblock A {B}ayesian nonparametric approach for time series clustering.
\newblock {\em Bayesian Analysis}, 9(1):147--170.

\bibitem[Roick et~al., 2019]{roick2019clustering}
Roick, T., Karlis, D., and McNicholas, P.~D. (2019).
\newblock Clustering discrete valued time series.
\newblock {\em arXiv preprint arXiv:1901.09249}.

\bibitem[Saad and Mansinghka, 2018]{saad2018temporally}
Saad, F. and Mansinghka, V. (2018).
\newblock Temporally-reweighted chinese restaurant process mixtures for
  clustering, imputing, and forecasting multivariate time series.
\newblock In {\em International Conference on Artificial Intelligence and
  Statistics}, pages 755--764.

\bibitem[Smith and Brown, 2003]{smith2003estimating}
Smith, A.~C. and Brown, E.~N. (2003).
\newblock Estimating a state-space model from point process observations.
\newblock {\em Neural computation}, 15(5):965--991.

\bibitem[St~Hilaire et~al., 2017]{st2017modeling}
St~Hilaire, M.~A., R{\"u}ger, M., Fratelli, F., Hull, J.~T., Phillips, A.~J.,
  and Lockley, S.~W. (2017).
\newblock Modeling neurocognitive decline and recovery during repeated cycles
  of extended sleep and chronic sleep deficiency.
\newblock {\em Sleep}, 40(1).

\bibitem[Vere-Jones, 2003]{vere2003introduction}
Vere-Jones, D. (2003).
\newblock {\em An Introduction to the Theory of Point Processes: Volume I:
  Elementary Theory and Methods}.
\newblock Springer.

\end{thebibliography}

\newpage 
\onecolumn
\appendix
\makeatletter
\newcommand{\algmargin}{\the\ALG@thistlm}   
\makeatother
\algnewcommand{\parState}[1]{\State \parbox[t]{\dimexpr\linewidth-\algmargin}{\strut #1\strut}}

\begin{center}
{\huge \textbf{APPENDICES}}
\end{center}

\section{MODEL EXTENSIONS} \label{model-extensions}

\subsection{Multidimensional Time Series}
Although this paper only considers examples in which each $\idx[t] x n \in \R$ and each $\idx[t] y n$ is one-dimensional, our model and inference algorithm can be extended to cases in which observed time series and/or latent states have multiple dimensions.   For example, as demonstrated in Section 5.2 of \citep{heng2017controlled}, cSMC scales well with a 64-dimensional vector time series model, suggesting that our proposed clustering approach with particle filtering is also applicable to multivariate series. 

\subsection{Finite Mixture of Time Series}
It is simple to convert our model into one in which the true number of clusters $K$ is known a priori.  Instead of using a Dirichlet process, we can simply use a Dirichlet($\bd \alpha$) distribution in which $\bd \alpha$ is a $K$-dimensional vector with each $\idx \alpha k > 0$ for $k = 1, \ldots, K$.  Then, we can modify Equation \ref{crp} as:
\begin{align*}
\bd q \given \bd \alpha &\sim \text{Dirichlet}(\bd \alpha), \\
\idx z 1, \ldots, \idx z N \given \bd q &\sim \text{Multinomial}(N, \bd q), \\
\bidx \theta k \given G &\sim G, & 1 \leq k \leq K,
\end{align*} 
where $\bd q$ is an intermediary variable that is easy to integrate over.  

The resultant inference algorithm is simpler.  The only necessary modification to Algorithm \ref{alg} is that, when sampling cluster assignments, there is no longer any need for an auxiliary integer parameter $m \geq 1$ to represent the infinite mixture.  Thus, Equation \ref{discrete} becomes 
\begin{align*}
p(\idx z n = k \given \idx Z {-n}, \bd \alpha) = \frac{\idx N k}{N - 1 + \idx \alpha k}, & & k = 1, \ldots, K,
\end{align*}
where $\idx N k$ is the number of cluster assignments equal to $k$ in $\idx Z {-n}$.  The process of sampling cluster parameters remains exactly the same as in the infinite mixture case.   

\section{CONTROLLED SEQUENTIAL MONTE CARLO} \label{csmc-app}

A key step in sampling both the cluster assignments and the cluster parameters of Algorithm \ref{alg} is computing the parameter likelihood $p(\bd y \given \bd \theta)$ for an observation vector $\bd{y} = y_1, \ldots, y_T$ and a given set of parameters $\bd \theta$.  

Recall the state-space model formulation:
\begin{align*}
x_1 \given \bd \theta &\sim h(x_1; \bd \theta), \\
x_t \given x_{t-1}, \bd \theta &\sim f(x_{t-1}, x_t; \bd \theta),  & &1 < t \leq T, \\
y_t \given x_t, \bd \theta &\sim g(x_t, y_t; \bd \theta), & & 1 \leq t \leq T.
\end{align*}

\subsection{Bootstrap Particle Filter}

The bootstrap particle filter (BPF) of \cite{doucet2001introduction} is based on a sequential importance sampling procedure that iteratively approximates each filtering distribution $p(x_t \given y_1, \ldots, y_t, \bd \theta)$ with a set of $S$ particles $\{x_t^1, \ldots, x_t^S\}$ so that
\begin{align*}
\hat{p}(\bd y \given \bd \theta) = \prod_{t=1}^T \left(\frac{1}{S}\sum_{s=1}^S g(x_t^s, y_t; \bd \theta)\right)
\end{align*}  
is an unbiased estimate of the parameter likelihood $p(\bd y \given \bd \theta)$.  Algorithm \ref{bpf} provides a review of this algorithm.

\begin{algorithm}
\caption{\texttt{BootstrapParticleFilter}($\bd y$, $\bd \theta$, $f$, $g$, $h$)} \label{bpf}
  \begin{algorithmic}[1]
     \For{$s = 1, \ldots, S$}
    	\parState{Sample $x_1^s \sim h(x_1; \bd \theta)$ and weight $w_1^s = g(x_1^s, y_1; \bd \theta)$.}
    \EndFor
    \parState{Normalize $\{w_1^s\}_{s=1}^S = \{w_1^s\}_{s=1}^S / \sum_{s=1}^S w_1^s$.}
    \For{$t = 2, \ldots, T$}
    	\For{$s = 1, \ldots, S$}
		\parState{Resample ancestor index $a \sim \text{Categorical}(w_{t-1}^1, \ldots, w_{t-1}^S)$.}
    		\parState{Sample $x_t^s \sim f(x^a_{t-1}, x_t; \bd \theta)$ and weight $w_t^s = g(x_t^s, y_t; \bd \theta)$.}
   	 \EndFor
	 \parState{Normalize $\{w_t^s\}_{s=1}^S = \{w_t^s\}_{s=1}^S / \sum_{s=1}^S w_t^s$.}
    \EndFor \\
    \Return{Particles $\{\{x_1^s\}_{s=1}^S, \ldots, \{x_T^s\}_{s=1}^S\}$}  
   \end{algorithmic}
\end{algorithm}

There are a variety of algorithms for the resampling step of Line 7.  We use the systematic resampling method.

\noindent A common problem with the BPF is that although its estimate of $p(\bd y \given \bd \theta)$ is unbiased, this approximation may have high variance for certain observation vectors $\bd y$.  The variance can be reduced at the price of increasing the number of particles, yet this often significantly increases computation time and is therefore unsatisfactory.  To remedy our problem, we follow the work of \cite{heng2017controlled} in using controlled sequential Monte Carlo (cSMC) as an alternative to the standard bootstrap particle filter. 

\subsection{Twisted Sequential Monte Carlo}
The basic idea of cSMC is to run several iterations of twisted sequential Monte Carlo, a process in which we redefine the model's state transition density $f$, initial prior $h$, and state-dependent likelihood $g$ in a way that allows the BPF to produce lower-variance estimates without changing the parameter likelihood $p(\bd y \given \bd \theta)$. See also \cite{guarniero2017iterated} for a different iterative approach. Using a \emph{policy} $\gamma = \{\gamma_1, \ldots, \gamma_T\}$ in which each $\gamma_t$ is a positive and bounded function, we define,
\begin{align*}
&h^\gamma(x_1; \bd \theta) = \frac{h(x_1; \bd \theta) \cdot \gamma_1(x_1)}{H^\gamma(\bd \theta)}, & & \\
&f^\gamma_t(x_{t-1}, x_{t}; \bd \theta) = \frac{f(x_{t-1}, x_t; \bd \theta) \cdot \gamma_t(x_t)}{F^\gamma_t(x_{t-1}; \bd \theta)}, & & 1 < t \leq T,
\end{align*}
where $H^\gamma(\bd \theta) = \int h(x_1; \bd \theta) \gamma_1(x_1) dx_1$ and $F^\gamma_t(x_{t-1}; \bd \theta) = \int f(x_{t-1}, x_t; \bd \theta)\gamma_t(x_t) dx_t$ are normalization terms for the probability densities $h^\gamma$ and $f^\gamma_t$, respectively.  To ensure that the parameter likelihood estimate $\hat{p}(\bd y \given \bd \theta)$ remains unbiased under the twisted model, we define the twisted state-dependent likelihoods $g_1^\gamma, \ldots, g_T^\gamma$ as functions that satisfy:
\begin{align*}
\hat{p}(\bd{x}, \bd{y} \given \bd \theta) &= h^\gamma(x_1; \bd \theta) \cdot \prod_{t=2}^T f^\gamma_t(x_{t-1}, x_t; \bd \theta) \cdot \prod_{t=1}^T g^\gamma_t(x_t, y_t; \bd \theta) \\
h(x_1; \bd \theta) \cdot \prod_{t=2}^T f(x_{t-1}, x_t; \bd \theta) \cdot \prod_{t=1}^T g(x_t, y_t; \bd \theta) &= \frac{h(x_1; \bd \theta) \gamma_1(x_1)}{H^\gamma(\bd \theta)} \cdot \prod_{t=2}^T \frac{f(x_{t-1}, x_t; \bd \theta)\gamma_t(x_t; \bd \theta) }{F^\gamma_t(x_{t-1}; \bd \theta)} \cdot \prod_{t=1}^T g^\gamma_t(x_t, y_t; \bd \theta) \\
\prod_{t=1}^T g(x_t, y_t; \bd \theta) &= \frac{\gamma_1(x_1)}{H^\gamma(\bd \theta)} \cdot \prod_{t=2}^T \frac{\gamma_t(x_t) }{F^\gamma_t(x_{t-1}; \bd \theta)} \cdot \prod_{t=1}^T g^\gamma_t(x_t, y_t; \bd \theta).
\end{align*}
This equality can be maintained if we define $g^\gamma_1, \ldots, g^\gamma_T$ as follows,
\begin{align*}
g^\gamma_1(x_1, y_1; \bd \theta) &= \frac{H^\gamma(\bd \theta) \cdot g(x_1, y_1; \bd \theta) \cdot F^\gamma_2(x_1; \bd \theta)}{\gamma_1(x_1)}, \\
g^\gamma_t(x_t, y_t; \bd \theta) &= \frac{g(x_t, y_t; \bd \theta) \cdot F^\gamma_{t+1}(x_t; \bd \theta)}{\gamma_t(x_t)}, & & 1 < t < T, \\
g^\gamma_T(x_T, y_T; \bd \theta) &= \frac{g(x_T, y_T; \bd \theta)}{\gamma_T(x_T)}.
\end{align*}
Thus, the parameter likelihood estimate of the twisted model is 
\begin{align*}
\hat{p}^\gamma(\bd y \given \bd \theta) = \prod_{t=1}^T \left(\frac{1}{S}\sum_{s=1}^S g^\gamma_t(x_t^s, y_t; \bd \theta)\right). 
\end{align*}  
The BPF is simply a degenerate case of twisted SMC in which $\gamma_t = 1$ for all $t$.

\subsection{Determining the Optimal Policy $\gamma^*$}
The variance of the estimate $\hat{p}^\gamma$ comes from the state-dependent likelihood $g$.  Thus, to minimize the variance, we would like $g^\gamma_t$ to be as uniform as possible with respect to $x_t$.  Let the optimal policy be denoted $\gamma^*$.  It follows that 
\begin{align*}
& \gamma^*_T(x_T) = g(x_T, y_T; \bd \theta), \\
& \gamma^*_t(x_t) = g(x_t, y_t; \bd \theta) \cdot F^{\gamma^*}_{t+1}(x_t; \bd \theta), & & 1 \leq t < T.
\end{align*}  
Under $\gamma^*$, the likelihood estimate $\hat{p}^{\gamma^*}(\bd y \given \bd \theta) = H^{\gamma^*} = p(\bd y \given \bd \theta)$ has zero variance.  However, it may be infeasible for us to use $\gamma^*$ in many cases, because the BPF algorithm requires us to sample $x_t$ from $f_t^{\gamma^*}$ for all $t$.  For example, under $\gamma^*$, we would have
\begin{align*}
f_T^{\gamma^*}(x_{T-1}, x_T; \bd \theta) \propto f(x_{T-1}, x_T; \bd \theta) \cdot \gamma^*_T(x_T) =  f(x_{T-1}, x_T; \bd \theta) \cdot g(x_T, y_T; \bd \theta), 
\end{align*}
which may be impossible to directly sample from if $f$ and $g$ form an intractable posterior (e.g. if $f$ is Gaussian and $g$ is binomial).  In such a case, we must choose a suboptimal policy $\gamma$.

\subsection{Choosing a Policy $\gamma$ for the Neuroscience Application}  
Recall the point-process state-space model (Equation \ref{bssm}), in which we have
\begin{align*} 
h(x_1; \bd \theta) &= \mathcal{N}(x_1 \given x_0 + \mu, \psi_0), \\
f(x_{t-1}, x_t ; \bd \theta) &= \mathcal{N}(x_t \given x_{t-1} , \psi),  \\
g(x_t, y_t) &= \text{Binomial}\left(M \cdot R, \frac{\exp x_t}{1 + \exp x_t}\right),
\end{align*} 
where we define the parameters $\bd \theta = \{\mu, \log \psi\}$ and $x_0, \psi_0, M, R$ are supplied constants.   

Here, we can show that $F^{\gamma^*}_{t+1}(x_t; \bd \theta) = \int f(x_{t}, x_{t+1}; \bd \theta)\gamma^*_{t+1}(x_{t+1}) dx_{t+1}$ must be log-concave in $x_{t}$.  This further implies that for all $t$, $\gamma^*_t(x_t) = g(x_t, y_t) \cdot F^{\gamma^*}_{t+1}(x_t; \bd \theta)$ is a log-concave function of $x_t$ since the product of two log-concave functions is log-concave.  Hence, we have shown that the optimal policy $\gamma^* = \{\gamma^*_1, \ldots, \gamma^*_T\}$ is a series of log-concave functions.  This justifies the approximation of each $\gamma^*_t(x_t)$ with a Gaussian function, 
\begin{align*}
\gamma_t(x_t) = \exp(-a_t x_t^2 - b_t x_t - c_t), & \quad (a_t, b_t, c_t) \in \R^3,
\end{align*}
and thus, $f_t^{\gamma}(x_{t-1}, x_t; \bd \theta) \propto f(x_{t-1}, x_t; \bd \theta) \cdot \gamma_t(x_t)$ is also a Gaussian density that is easy to sample from when running the BPF algorithm.

We want to find the values of $(a_t, b_t, c_t)$ that enforce $\gamma_t \approx \gamma_t^*$ for all $t$.  One simple way to accomplish this goal is to find the $(a_t, b_t, c_t)$ that minimizes the least-squares difference between $\gamma_t$ and $\gamma_t^*$ in log-space.  That is, given a set of samples $\{x_t^1, \ldots, x_t^S\}$ for the random variable $x_t$, we solve for: 
\begin{align*}
(a_t, b_t, c_t) &= \arg \min_{(a_t, b_t, c_t) \in \R^3} \sum_{s=1}^S \left[\log \gamma_t(x_t^s) - \log \gamma^*_t(x_t^s) \right]^2  \\
&=  \arg \min_{(a_t, b_t, c_t) \in \R^3} \sum_{s=1}^S \left[-(a_t (x_t^s)^2 + b_t (x_t^s) + c_t) - \log \gamma^*_t(x_t^s) \right]^2.
\end{align*}
Also note that in a slight abuse of notation, we redefine for all $t < T$,
\begin{align*}
\gamma_t^*(x_t) = g(x_t, y_t) \cdot F^\gamma_{t+1}(x_t; \bd \theta), 
\end{align*}
because when performing approximate backwards recursion, it is not possible to analytically solve for the intractable integral $F^{\gamma*}_{t+1}(x_t; \bd \theta)$.    

In the aforementioned least-squares optimization problem, there is one additional constraint that we must take into account.  Recall that $f_t^{\gamma}(x_{t-1}, x_t; \bd \theta) \propto f(x_{t-1}, x_t; \bd \theta) \cdot \gamma_t(x_t)$ is a Gaussian pdf that we sample from.  Therefore, we must ensure that the variance of this distribution is positive, which places a constraint on $\gamma_t$ and more specifically, the domain of $(a_t, b_t, c_t)$.  Using properties of Gaussians, we can perform algebraic manipulation to work out the following parameterizations of $h^\gamma$ and $f_t^\gamma$:
\begin{align*}
& h^\gamma(x_1; \bd \theta) = \Norm\left(x_1 \given[\Big] \frac{\psi_0^{-1} \cdot (x_0 + \mu) -b_1}{\psi_0^{-1} + 2a_1}, \frac{1}{\psi_0^{-1} + 2a_1}\right), \\
& f_t^\gamma(x_{t-1}, x_t; \bd \theta) = \Norm\left(x_t \given[\Big] \frac{\psi^{-1} \cdot x_{t-1} - b_t}{\psi^{-1} + 2a_t}, \frac{1}{\psi^{-1} + 2a_t}\right), & &1 < t \leq T.
\end{align*}

\noindent The corresponding normalizing terms for these densities are 
\begin{align*}
& H^\gamma(\bd \theta) = \frac{1}{\sqrt{1 + 2a_1 \psi_0}} \exp\left(\frac{\psi_0^{-1} \cdot (x_0 + \mu)  - (b_1)^2}{2(\psi_0^{-1} + 2a_1)} - \frac{ (x_0 + \mu)^2}{2 \psi_0} - c_1\right), \\
& F^\gamma_{t}(x_{t-1}; \bd \theta) =  \frac{1}{\sqrt{1 + 2a_t \psi}} \exp\left(\frac{\psi^{-1} \cdot x_{t-1}  - (b_t)^2}{2(\psi^{-1} + 2 a_t)} - \frac{x_{t-1}^2}{2\psi} - c_t\right), & & 1 < t \leq T.
\end{align*}

\noindent Thus, to obtain $(a_t, b_t, c_t)$ and consequently $\gamma_t$ for all $t$, we solve the aforementioned least-squares minimization problem subject to the following constraints:
\begin{align*}
a_1 > - \frac{1}{2\psi_0}, & & a_t > -\frac{1}{2\psi}, \quad 1 < t \leq T.
\end{align*}

\subsection{Full cSMC Algorithm}
The full controlled sequential Monte Carlo algorithm iterates on twisted SMC for $L$ iterations, building a series of policies $\aidx \gamma 1, \aidx \gamma 2, \ldots, \aidx \gamma L$  over time.    
Given two policies $\Gamma'$ and $\gamma$, we can define
\begin{align*}
&h^{\Gamma' \cdot \gamma}(x_1) \propto h^{\Gamma'}(x_1) \gamma_1(x_1) = h(x_1; \bd \theta) \cdot {\Gamma'_1}(x_1) \cdot \gamma_1(x_1), & \\ 
& f^{\Gamma' \cdot \gamma}_t(x_{t-1}, x_{t}; \bd \theta) \propto f^{\Gamma'}_t(x_{t-1}, x_t; \bd \theta) \cdot \gamma_t(x_t) = f(x_{t-1}, x_t; \bd \theta) \cdot \Gamma'_t(x_t) \cdot \gamma_t(x_t), & & 1 < t \leq T.
\end{align*}
We can see from these relationships that twisting the original model using $\Gamma'$ and then twisting the new model using $\gamma$ has the same effect as twisting the original model using a cumulative policy $\Gamma$ where each $\Gamma_t(x_t) = \Gamma'_t(x_t) \cdot \gamma_t(x_t)$.  We state the full cSMC algorithm in Algorithm \ref{csmc}.

\begin{algorithm}[H]
\caption{\texttt{ControlledSMC}($\bd y$, $g$, $\mu$, $\psi$, $x_0$, $\psi_0$, $L$)} \label{csmc}
  \begin{algorithmic}[1]
    \parState{Define $f(x_{t-1}, x_t; \bd \theta) = \Norm(x_t \given x_{t-1}, \psi)$ and $h(x_1; \bd \theta) = \Norm(x_1 \given x_0 + \mu, \psi_0)$.} 
    \parState{Define parameters $\bd \theta = \{\mu, \log \psi\}$.}
    \parState{Collect particles $\{x_1^s\}_{s=1}^S, \ldots, \{x_T^s\}_{s=1}^S$ from \texttt{BootstrapParticleFilter}($\bd y$, $\bd \theta$, $f$, $g$, $h$).}
    \parState{Initialize $\Gamma' = \{\Gamma'_1, \ldots, \Gamma'_T\}$ where $\Gamma'_t(x_t) = 1$ for all $t = 1, \ldots, T$.}
    \parState{Initialize $g^{\Gamma'}_t(x_t, y_t) = g(x_t, y_t)$ for all $t = 1, \ldots, T$.}
    \parState{Initialize $\aidx[t] a 0 = 0, \aidx[t] b 0 = 0, \aidx[t] c 0 = 0$ for all $t = 1, \ldots, T$.}
    
    \For{$\ell =1, \ldots, L$}
        \item[\quad \quad // \emph{Approximate backward recursion to determine policy and associated functions}]
    \parState{Define $\gamma^*_T(x_T) = g_T^{\Gamma'} (x_T, y_T).$}
    \For{$t = T, \ldots, 2$}
    	\parState{Solve $(\aidx[t] a \ell, \aidx[t] b \ell, \aidx[t] c \ell) = \arg \min_{(a_t, b_t, c_t)} \sum_{s=1}^S \left[-(a_t (x_t^s)^2 + b_t (x_t^s) + c_t) - \log \gamma^*_t(x_t^s)\right]^2$ subject to $a_t > -1 / (2 \psi) -  \sum_{\ell'=0}^{\ell-1} \aidx[t] a {\ell'}$ using linear regression.}
	\parState{Define new policy function $\gamma_t (x_t) = \exp(-\aidx[t] a \ell x_t^2 - \aidx[t] b \ell x_t - \aidx[t] c \ell)$.} 
	\parState{Define cumulative policy function $\Gamma_t(x_t) = \Gamma'_t(x_t) \cdot \gamma_t(x_t) = \exp(-A_t x_t^2 - B_t x_t - C_t)$ where $A_t = \sum_{\ell'=0}^\ell \aidx[t] a {\ell'}$, $B_t = \sum_{\ell'=0}^\ell \aidx[t] b {\ell'}$, and $C_t = \sum_{\ell'=0}^\ell \aidx[t] c {\ell'}$.}
	\parState{Define $f^\Gamma_t (x_{t-1}, x_t; \bd \theta)$ and $F^\Gamma_{t}(x_{t-1}; \bd \theta)$.}
	\If{$t = T$}
		\parState{Define $g^\Gamma_T(x_T, y_T) = g(x_T, y_T) / \Gamma_T(x_T)$.}
	\Else 
		\parState{Define $g^\Gamma_t(x_t, y_t) = g(x_t, y_t) \cdot F^\Gamma_{t+1}(x_t; \bd \theta) / \Gamma_t(x_t)$.}
	\EndIf
	\parState{Define $\gamma^*_{t-1}(x_{t-1}) = g^{\Gamma'}_{t-1}(x_{t-1}, y_{t-1}) \cdot F^{\Gamma}_{t}(x_{t-1}; \bd \theta) / F_t^{\Gamma'}(x_{t-1}; \bd \theta)$.}	
    \EndFor
    \parState{Solve $(\aidx[1] a \ell, \aidx[1] b \ell, \aidx[1] c \ell) = \arg \min_{(a_1, b_1, c_1)} \sum_{s=1}^S \left[-(a_1 (x_1^s)^2 + b_1 (x_1^s) + c_1) - \log \gamma^*_1(x_1^s)\right]^2$ subject to $a_1 > -1 / (2 \psi_0) -  \sum_{\ell'=0}^{\ell-1} \aidx[1] a {\ell'}$ using linear regression.}
    \parState{Define new policy function $\gamma_1(x_1) = \exp(-\aidx[1] a \ell x_1^2 - \aidx[1] b \ell x_1 -\aidx[1] c \ell).$}
    \parState{Define cumulative policy function $\Gamma_1(x_1) = \Gamma'_1(x_1) \cdot \gamma_1(x_1) = \exp(-A_1 x_t^2 - B_1 x_t - C_1)$ where $A_1 = \sum_{\ell'=0}^\ell \aidx[1] a {\ell'}$, $B_1 = \sum_{\ell'=0}^\ell \aidx[1] b {\ell'}$, and $C_1 = \sum_{\ell'=0}^\ell \aidx[1] c {\ell'}$.}
    \parState{Define $\Gamma$-twisted initial prior $h^\Gamma(x_1; \bd \theta)$ and $H^\Gamma(\bd \theta)$.}
    \parState{Define $g^\Gamma_1(x_1, y_1) = H^\Gamma(\bd \theta) \cdot g(x_1, y_1) \cdot F^\Gamma_2(x_1; \bd \theta) / \Gamma_1(x_1)$.}
    \item[\quad \quad // \emph{Forward bootstrap particle filter to sample particles and compute weights}]
    \For{$s = 1, \ldots, S$}
    	\parState{Sample $x_1^s \sim h^\Gamma(x_1)$ and weight $w_1^s = g^\Gamma_1(x_1^s, y_1)$.}
    \EndFor
    \parState{Normalize $\{w_1^s\}_{s=1}^S = \{w_1^s\}_{s=1}^S / \sum_{s=1}^S w_1^s$.}
    \For{$t = 2, \ldots, T$}
    	\For{$s = 1, \ldots, S$}
		\parState{Resample ancestor index $a \sim$ Categorical$(w_{t-1}^1, \ldots, w_{t-1}^S)$.}
    		\parState{Sample $x_t^s \sim f_t^\Gamma(x^a_{t-1}, x_t; \bd \theta)$ and weight $w_t^s = g^\Gamma_t(x_t^s, y_t)$.}
   	 \EndFor
	 \parState{Normalize $\{w_t^s\}_{s=1}^S = \{w_t^s\}_{s=1}^S / \sum_{s=1}^S w_t^s$.}
    \EndFor
    \parState{Update $\Gamma' = \Gamma$.}
    \EndFor \\
    \Return{Likelihood estimate $\hat{p}^\Gamma(\bd{y} \given\bd \theta)$.}  
   \end{algorithmic}
\end{algorithm}

\section{MULTIPLE GIBBS SAMPLES WITH OPTIMAL CO-OCCURENCE MATRIX} \label{mult-co}
This section extends the method of selecting clusters detailed in Section \ref{ssec:sel-clust}.  After running the DPnSSM inference algorithm (Algorithm \ref{alg}), we construct co-occurrence matrices $\baidx \Omega i$  for $i = 1, \ldots, I$ (Equation \ref{co-occurrence}).  Then, we select the optimal Gibbs sample $i^*$.  If there are $J > 1$ Gibbs samples $i_1, \ldots, i_J$ such that $\baidx \Omega {i_j} = \baidx \Omega {i^*}$ for $j = 1, \ldots, J$, our final cluster parameters $\bd \Theta^*$ can be redefined as the average among the corresponding parameter samples,
\begin{align*} \label{final-params}
\bd \Theta^* = \frac{1}{J} \sum_{j=1}^J \baidx \Theta {i_j} \approx \E[\bd \Theta \given Z = Z^*].
\end{align*}
This averaging must be preceded by a permutation of each set of $\bidx \theta 1, \ldots, \bidx \theta K \in \baidx \Theta {i_j^*}$ to fix any potential label switching.     

\section{ROBUSTNESS OF MODEL TO STIMULUS MISSPECIFICATION} \label{sim-robust}
\begin{figure}[h]
\begin{center}
\includegraphics[scale=0.26]{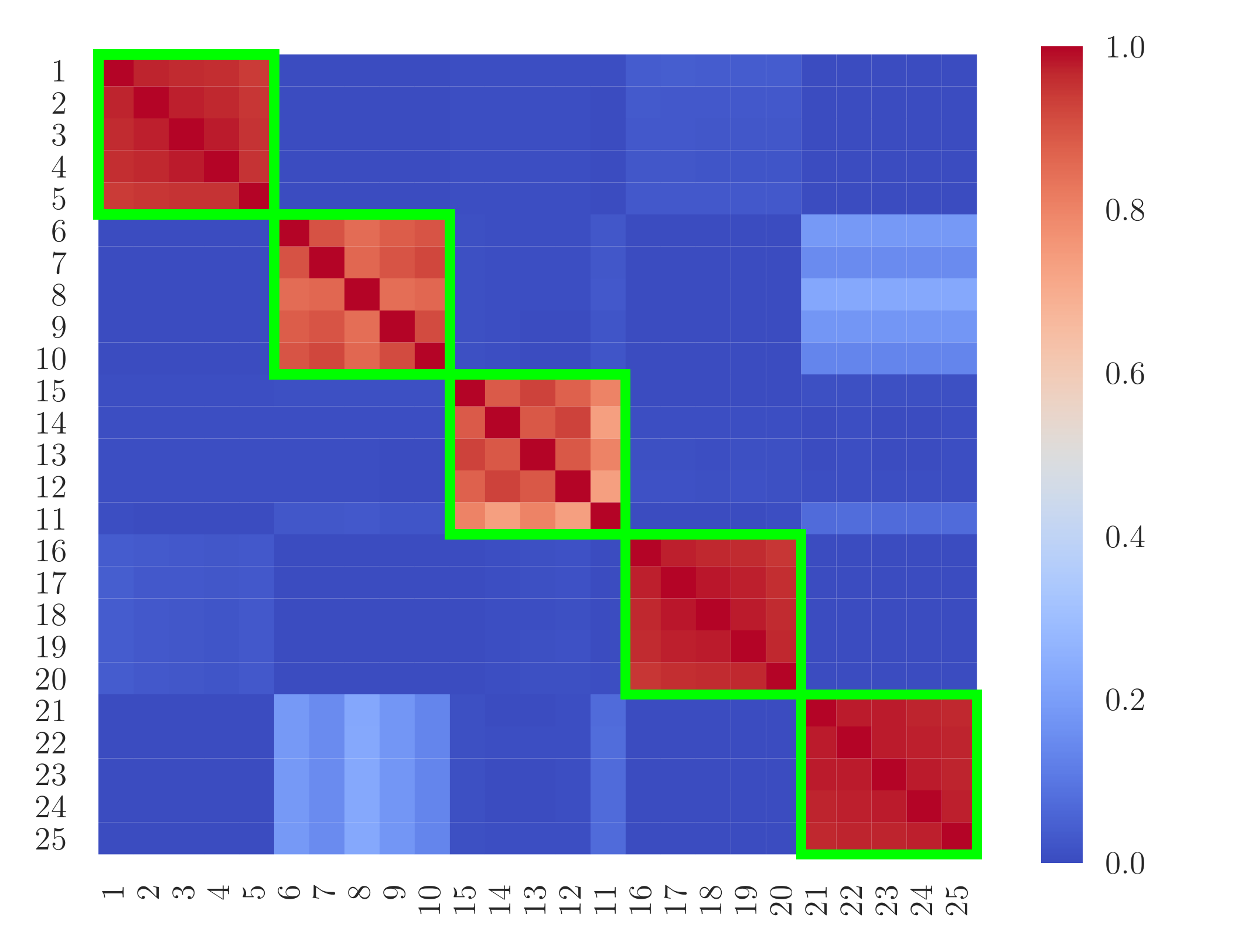}
\includegraphics[scale=0.26]{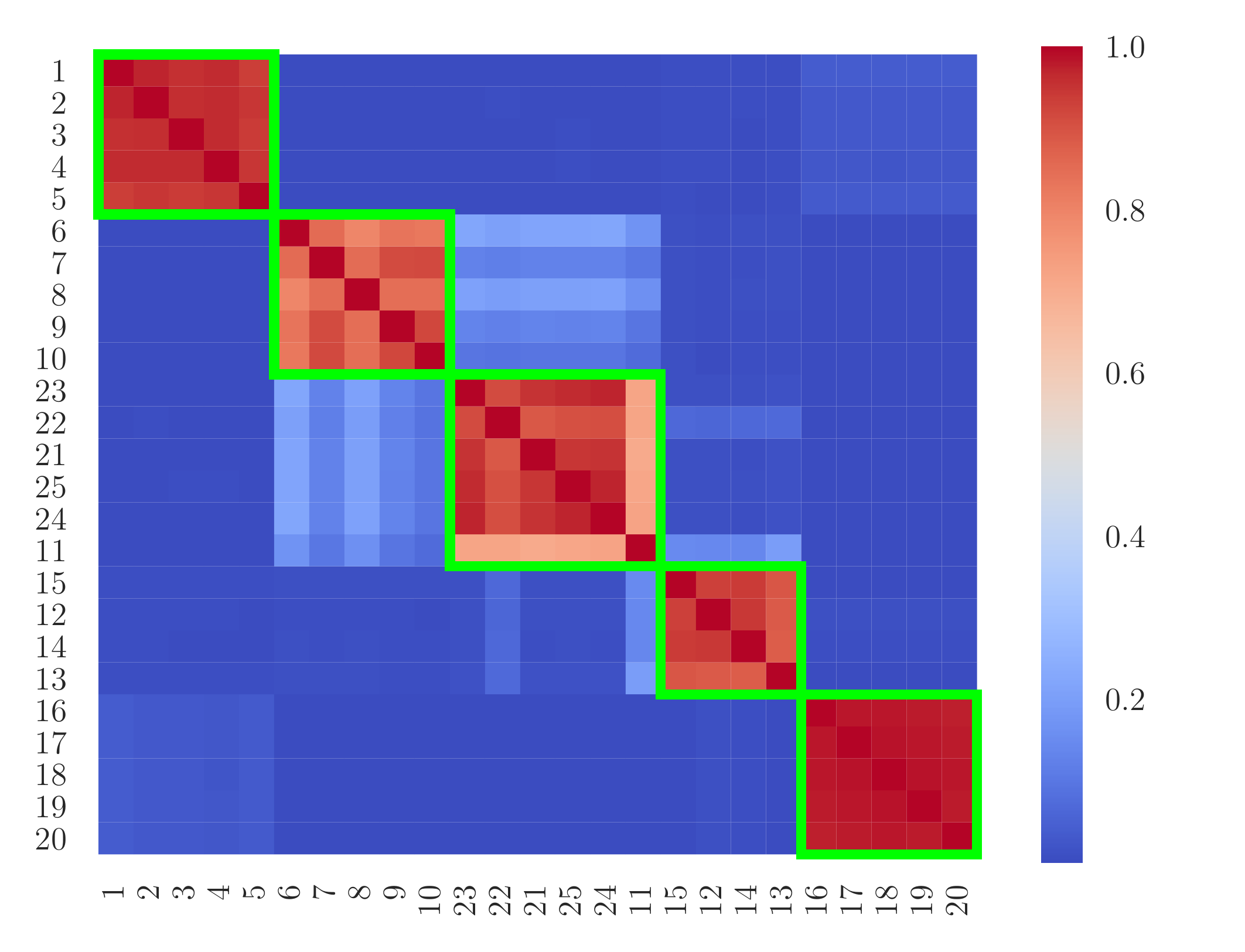}
\includegraphics[scale=0.26]{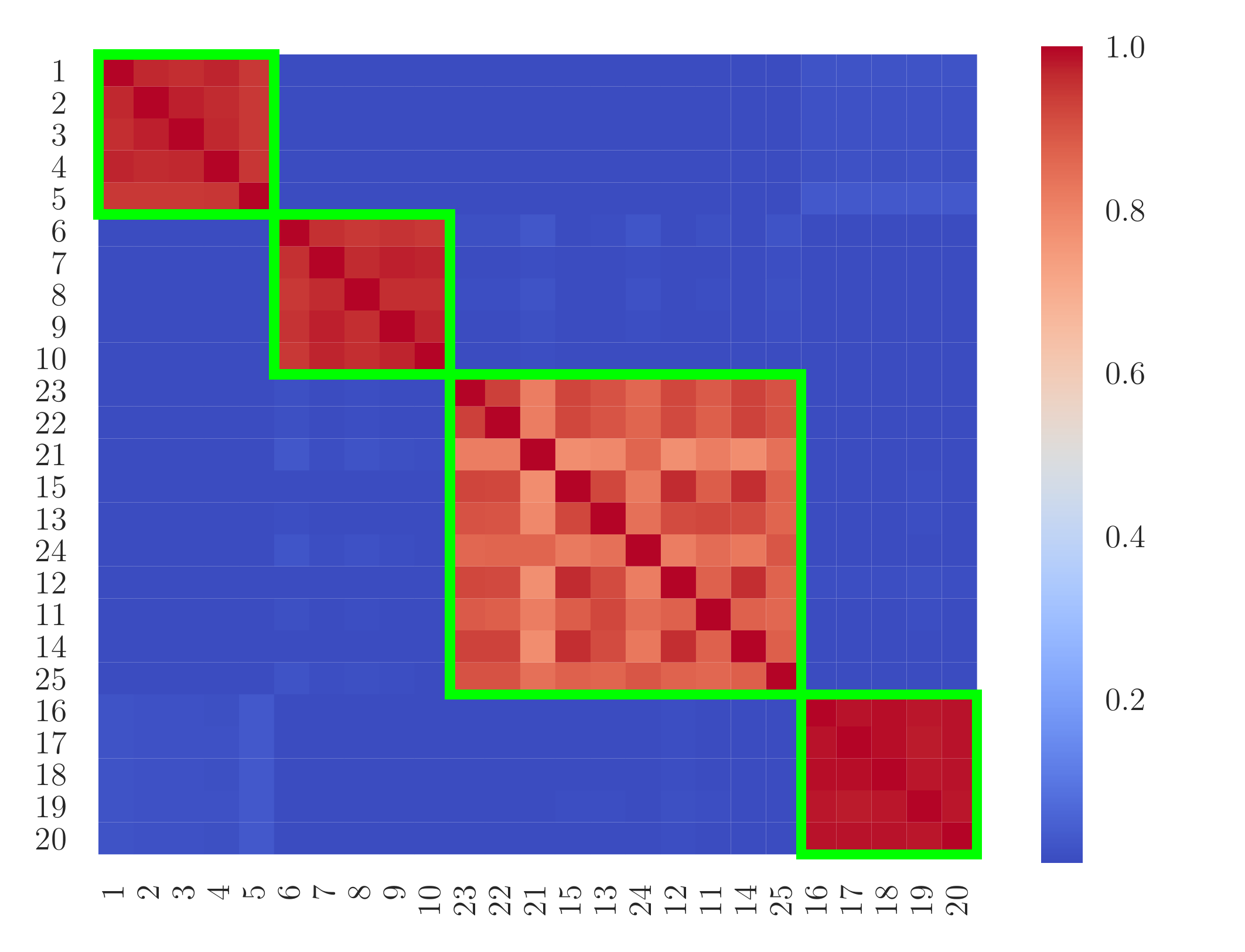}
\end{center}
\caption{Results from running the DPnSSM inference algorithm in cases in which the model specifies the stimulus as occurring (left) 40 ms, (center) 80 ms, and (right) 160 ms \emph{after} the true onset.  At 40 ms, the ground-truth clustering can be recovered, but this ability decays as the time difference increases.} \label{overpredict}
\end{figure}

We extend the results of Section \ref{ssec:sim-data} by testing the robustness of the model under cases in which there is a mismatch between the true stimulus onset and the model's specification of the stimulus onset.  In particular, we first examine the case in which the model overpredicts the true stimulus onset.  Figure \ref{overpredict} presents heatmaps of the mean co-occurrence matrix $\bd \Omega$ in cases in which the model's anticipation of the stimulus falls 40 ms, 80 ms, and 160 ms behind the true onset.  Table \ref{robust-table1} lists the parameters chosen in the final clustering of the data in these three cases.  In all experiments, we use the same data generation process as detailed in Section \ref{sssec:sim-data} and the same modeling process as detailed in Section \ref{sim-model-details}.  

\begin{table}[h]
\vspace{-3mm}
\caption{Final cluster parameters for model stimulus delay.  As expected, the absolute value of $\idx {\mu^*} k$ decreases for all $k$ as time mismatch increases.} \label{robust-table1}
\begin{center}
\begin{tabular}{c|ccc|ccc|cccl}
\multicolumn{1}{c}{} &\multicolumn{3}{c}{40 ms model delay} &\multicolumn{3}{c}{80 ms model delay} &\multicolumn{3}{c}{160 ms model delay} \\
\hline
$k$ & $\idx {\mu^*} k$ & $\log \idx {\psi^*} k$ &\# of Neurons & $\idx {\mu^*} k$ & $\log \idx {\psi^*} k$ &\# of Neurons & $\idx {\mu^*} k$ & $\log \idx {\psi^*} k$ &\# of Neurons \\
\hline
$1$ & $+0.86$ & $-10.86$ & 5 & $+0.80$ & $-10.95$ & 5 & $+0.65$ & $-10.79$ & 5  \\
$2$ &$-0.88$ & $-12.12$ & 5 & $-0.84$ & $-12.45$ & 5 & $-0.76$ & $-12.48$ & 5 \\
$3$ &$+0.02$ & $-10.31$  & 5 & $-0.62$ & $-6.23$ & 6 & $+0.01$ & $-9.36$ & 10 \\
$4$ &$+0.94$ & $-5.53$ & 5 & $+0.05$ & $-10.77$ & 4 & $+0.57$ & $-5.47$ & 5 \\
$5$ &$-0.87$ & $-5.91$ & 5& $+0.87$ & $-5.52$ & 5 
\end{tabular}
\vspace{-4mm}
\end{center}\end{table}   

\begin{figure}[h]
\begin{center}
\includegraphics[scale=0.26]{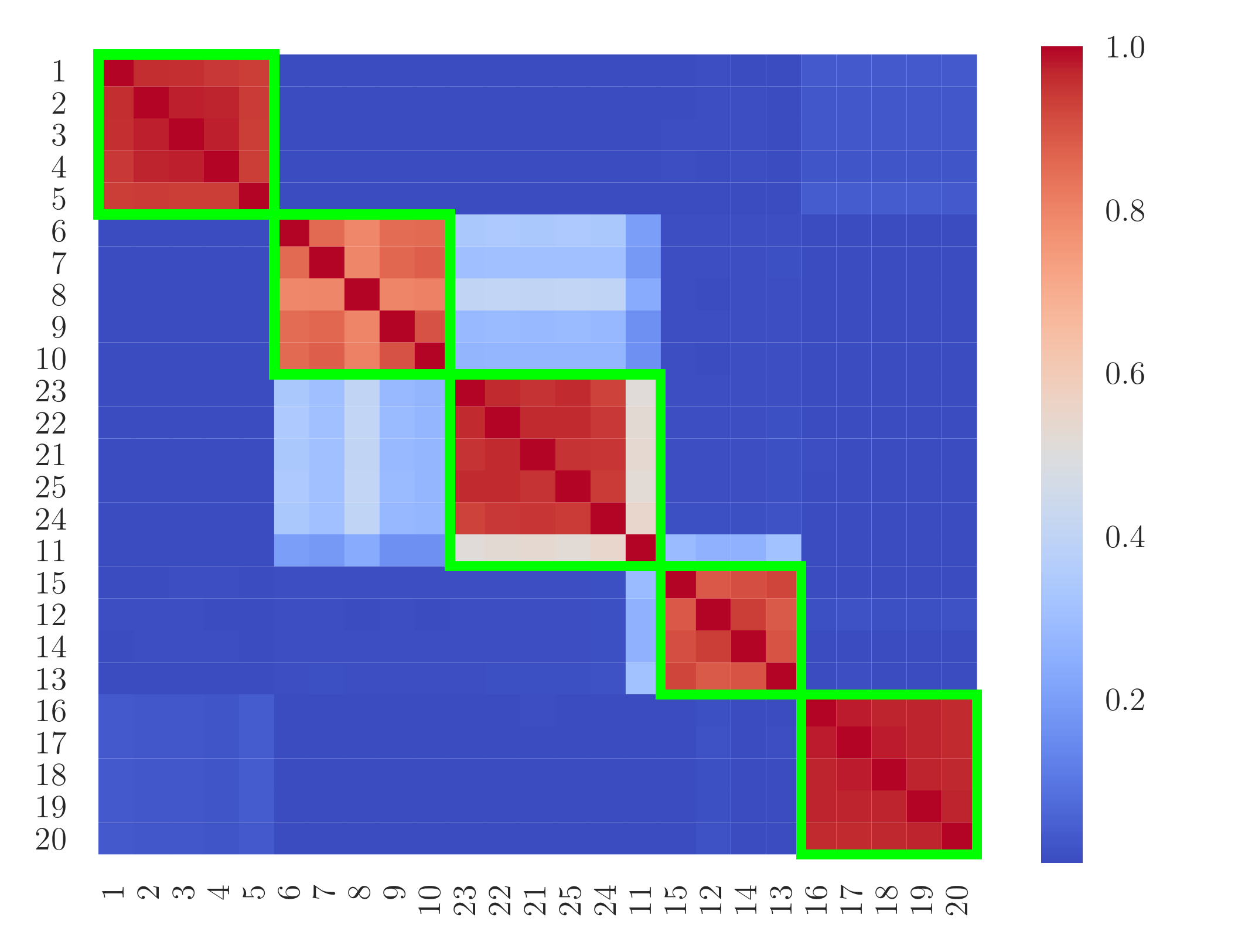}
\includegraphics[scale=0.26]{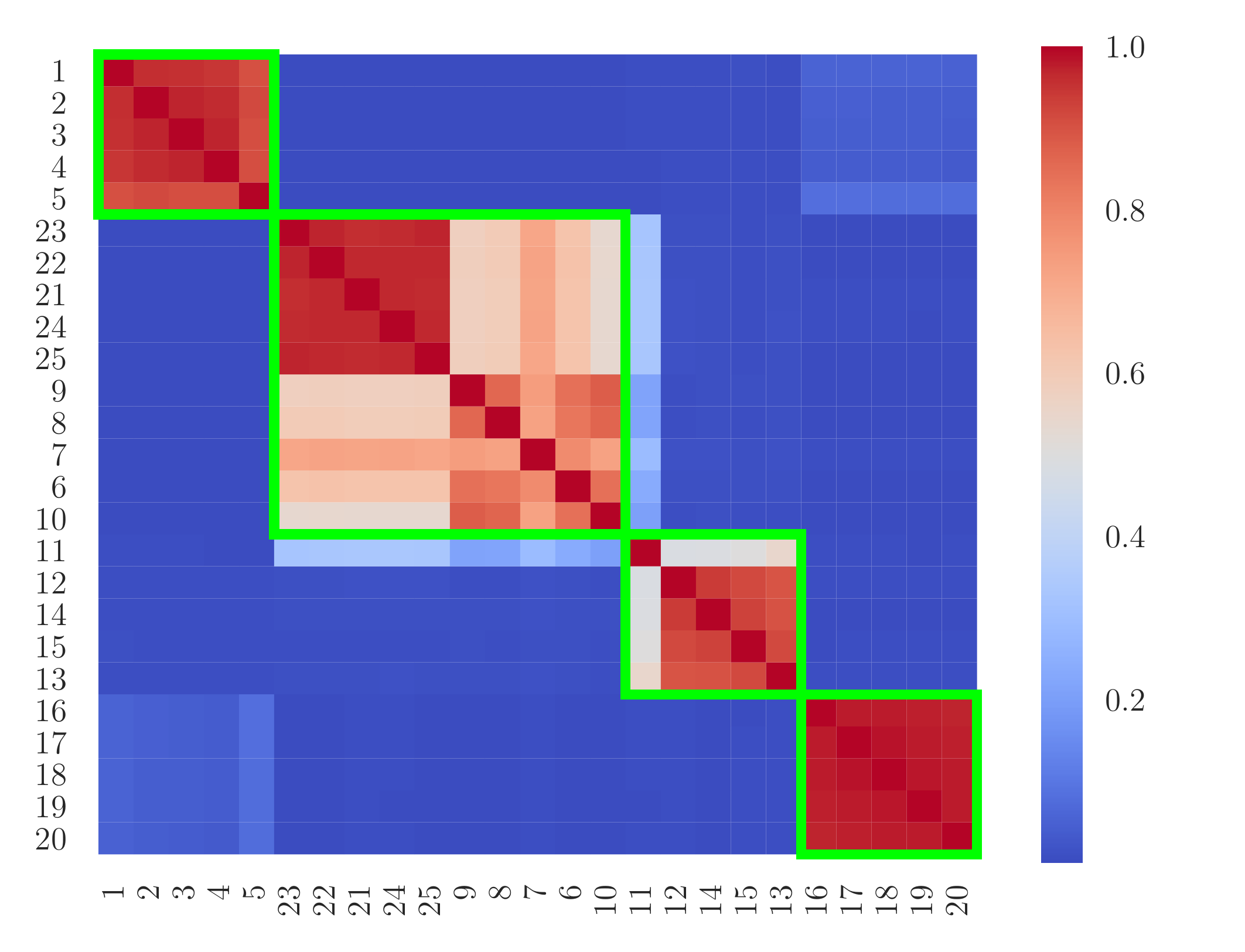}
\includegraphics[scale=0.26]{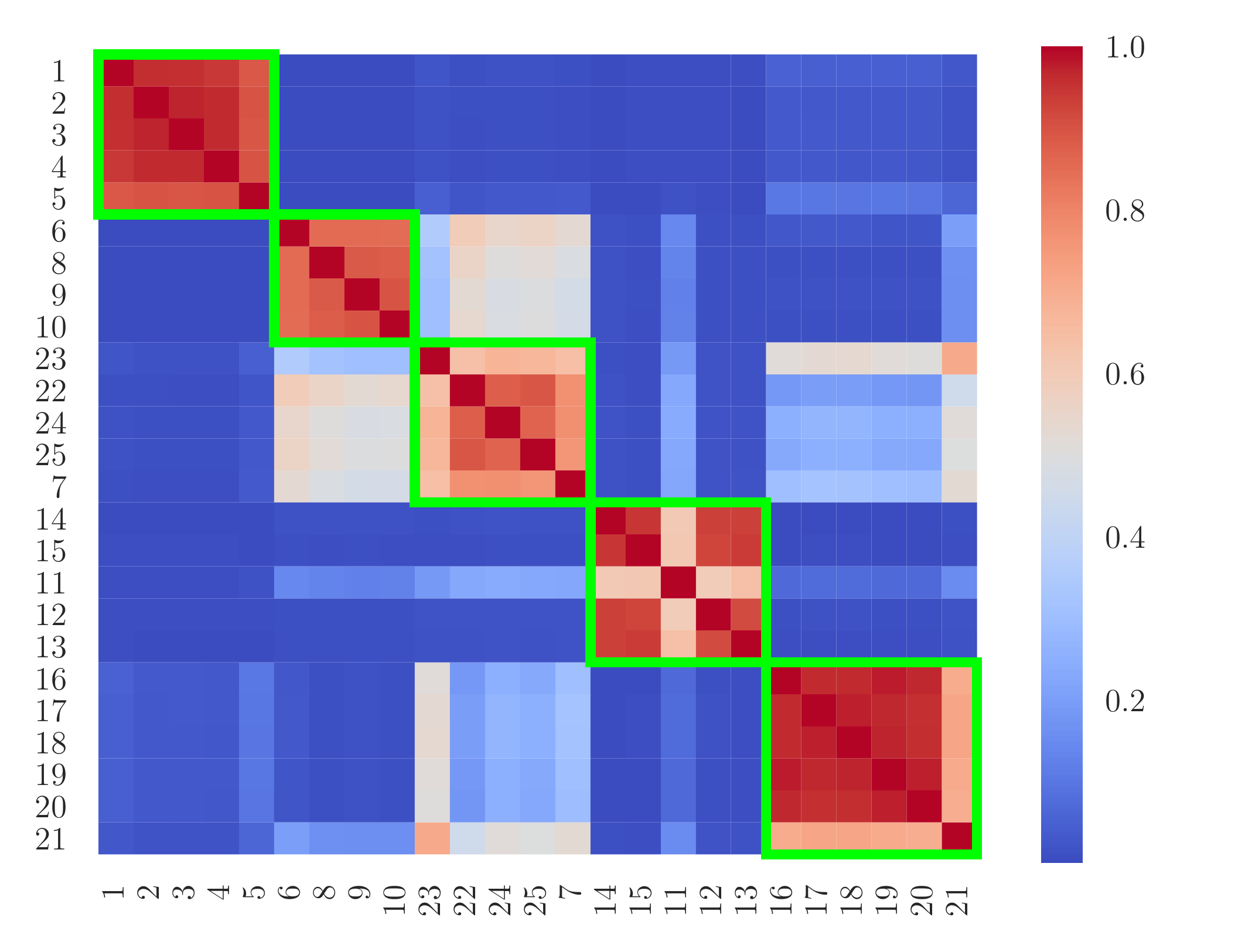}
\end{center}
\caption{Results from running the DPnSSM inference algorithm in cases in which the model specifies the stimulus as occurring (left) 10 ms, (center) 20 ms, and (right) 40 ms \emph{before} the true onset.  At 10 ms, the ground-truth clustering can almost be fully recovered, but this ability significantly decays as the time difference increases.  For a true stimulus delay of 40 ms, there even exists one cluster containing both excited and inhibited neurons.} \label{underpredict}
\end{figure}

Next, we examine the case in which the model underpredicts the true stimulus onset -- by 10 ms, 20 ms, and 40 ms.  This set of results is less robust than the previous one.  Heatmaps are given in Figure \ref{underpredict}, while parameters are given in Table \ref{robust-table2}.  

\begin{table}[h]
\vspace{-3mm}
\caption{Final cluster parameters for true stimulus delay.} \label{robust-table2}
\begin{center}
\begin{tabular}{c|ccc|ccc|cccl}
\multicolumn{1}{c}{} &\multicolumn{3}{c}{10 ms stimulus delay} &\multicolumn{3}{c}{20 ms stimulus delay} &\multicolumn{3}{c}{40 ms stimulus delay} \\
\hline
$k$ & $\idx {\mu^*} k$ & $\log \idx {\psi^*} k$ &\# of Neurons & $\idx {\mu^*} k$ & $\log \idx {\psi^*} k$ &\# of Neurons & $\idx {\mu^*} k$ & $\log \idx {\psi^*} k$ &\# of Neurons \\
\hline
$1$ & $+0.93$ & $-10.45$ & 5 & $+0.91$ & $-10.37$ & 5 & $+0.79$ & $-8.71$ & 5  \\
$2$ &$-0.90$ & $-12.47$ & 4 & $-0.63$ & $-6.52$ & 10 & $-0.89$ & $-11.5$ & 4 \\
$3$ &$-0.66$ & $-6.08$  & 6 & $+0.03$ & $-10.02$ & 5 & $-0.34$ & $-5.93$ & 5 \\
$4$ &$+0.07$ & $-10.96$ & 5 & $+0.67$ & $-5.31$ & 5 & $+0.02$ & $-10.41$ & 5 \\
$5$ &$+0.82$ & $-5.44$ & 5 & & & & $+0.34$ & $-4.96$ & 6
\end{tabular}
\vspace{-4mm}
\end{center}\end{table}

\section{ADDITIONAL RASTER PLOTS FOR CUE DATA OVER TIME} \label{raster-plots}
\begin{figure}[H]
\begin{center}
\includegraphics[scale=0.5]{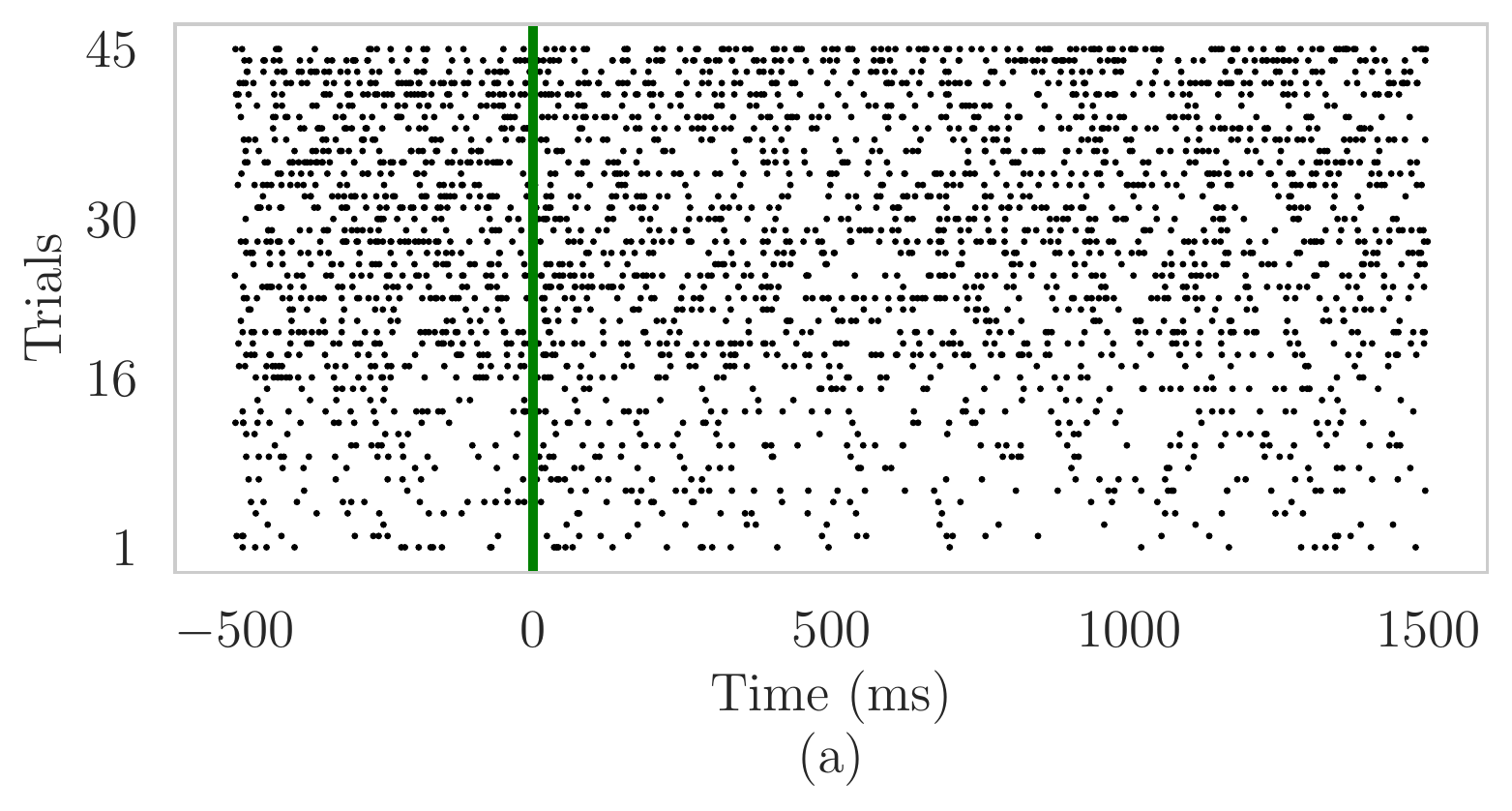}
\includegraphics[scale=0.5]{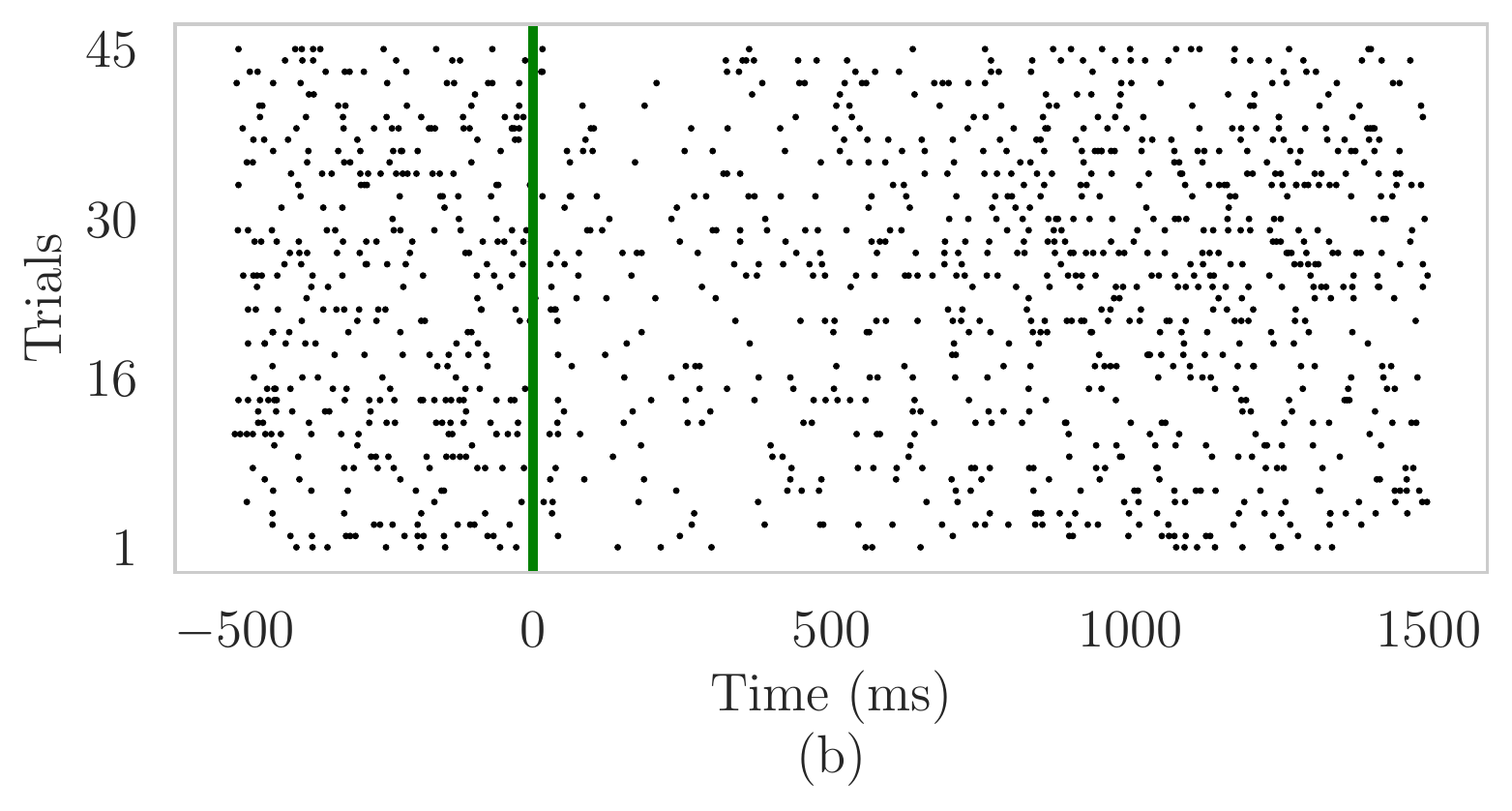}
\includegraphics[scale=0.5]{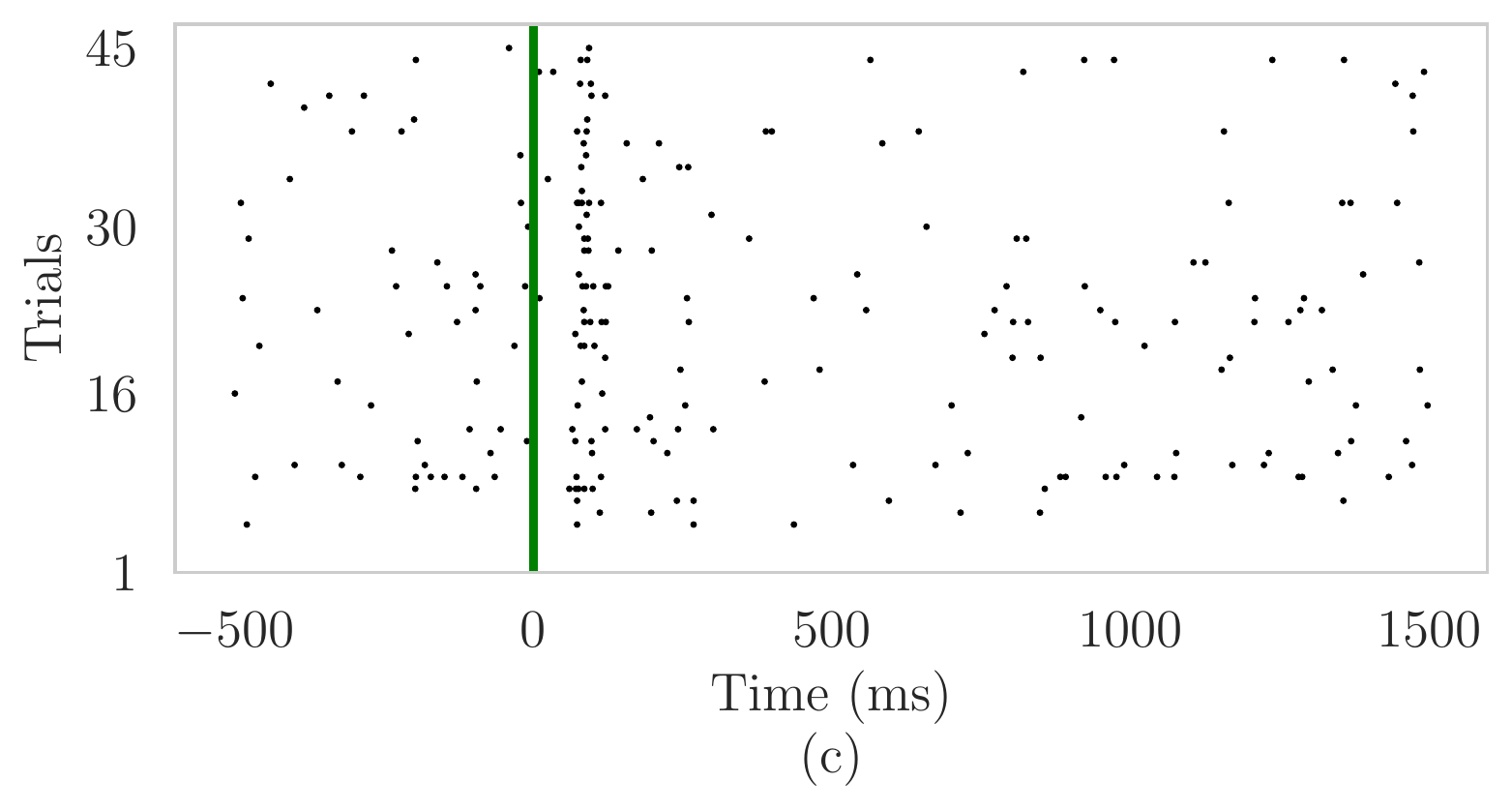}
\end{center}
\vspace{-4mm}
\caption{Overlaid raster plots of additional neuronal clusters found in Section \ref{cue-cluster} for (a) cluster $k =2$ with eight neurons and slightly inhibited/sustained responses, (b) cluster $k = 4$ with four neurons and more strongly inhibited/variable responses, and (c) cluster $k = 2$ with a single neuron and a delayed excited effect.  A black dot at $(\tau, r)$ indicates a spike from one of the neurons in the corresponding cluster at time $\tau$ during trial $r$. The vertical green line indicates cue onset.} \label{cue-rasters-app}
\end{figure}

\section{DETAILS OF CSMC VS. BPF EXPERIMENT} \label{csmc-bpf-app}
This section describes the experimental setup of Section \ref{ssec:bpf-csmc} that is used to produce Figure \ref{bpf-csmc}.  Computational cost is fixed at approximately 35 milliseconds per likelihood evaluation for either method.  For the bootstrap particle filter (BPF), we use $S = 1024$ particles.  For controlled sequential Monte Carlo (cSMC), we use $S = 64$ particles and $L = 3$ iterations.  All parameter log-likelihood evaluations are performed on a representative real neuron's cue data over time $\bd y$, as described in Section \ref{sssec:time-data}.  For each particle filtering method, let $v(\mu, \psi) = \Var[\log p(\bd y \given \bd \theta)]$, where $\bd \theta = [\mu, \log \psi]^\top$.  Figure \ref{bpf-csmc} plots empirical estimates of $\hat{v}$ of $v$ over different values of $(\mu, \psi)$ for the two methods.  For each empirical variance estimate, we use 500 estimates of $\log p(\bd y \given \bd \theta)$.  As $\log \psi$ decreases, cSMC performs substantially better than BPF, especially for extreme values of $\mu$.

\end{document}